\documentclass{article}

    \PassOptionsToPackage{numbers, compress}{natbib}
 \usepackage[preprint]{neurips_2026}

\usepackage[utf8]{inputenc} 
\usepackage[T1]{fontenc}    
\usepackage{hyperref}       
\usepackage{url}            
\usepackage{booktabs}       
\usepackage{amsfonts}       
\usepackage{nicefrac}       
\usepackage{microtype}      
\usepackage{xcolor}         
\let\OriginalAddContentsLine\addcontentsline

\usepackage{graphicx}
\usepackage{amsmath}
\usepackage{amssymb}
\usepackage{mathtools}
\usepackage{amsthm}
\usepackage[most]{tcolorbox} 
\usepackage{wrapfig}
\usepackage{pifont}
\usepackage[textsize=tiny]{todonotes}
\usepackage{multirow}
\usepackage{float}
\usepackage{subcaption}
\usepackage{colortbl}
\usepackage{tocloft}
\usepackage{etoc}

\theoremstyle{plain}
\newtheorem{theorem}{Theorem}[section]
\newtheorem{proposition}[theorem]{Proposition}

\theoremstyle{definition}
\newtheorem{definition}[theorem]{Definition}

\theoremstyle{remark}

\newcommand{\nosection}[1]{\vspace{2pt}\noindent\textbf{#1.}}

\newcommand{\modelname}{ASAM}

\title{Beyond Binary Edits: Robust Multimodal Knowledge \\ Editing with Adversarial Subspace Alignment}

\author{
Haoyuan Wang$^{1}$ \ \ \ \ \
Xiaohao Liu$^{2}$\ \ \ \ \ 
Jiajie Su$^{1}$ \ \ \ \ \
Jianmao Xiao$^{3}$ \ \ \ \ \ 
Chaochao Chen$^{1}$ \ \ \ \ \\\
$^1$Zhejiang University\quad \quad$^2$National University of Singapore \quad \quad$^3$Jiangxi Normal University\\
\small \texttt{\{haoyuanwang, sujiajie, zjuccc\}@zju.edu.cn,} \\
\small \texttt{xiaohao.liu@u.nus.edu}  \\
}

\begin{document}

\maketitle

\begin{abstract}
Multimodal large language models (MLLMs) need efficient mechanisms to update knowledge without degrading existing capabilities. While intrinsic multimodal knowledge editing achieves strong reliability and locality, it often exhibits limited generality, failing to propagate edits across semantically equivalent visual and linguistic variations. This issue arises from the lack of explicit semantic supervision, rigid editing scopes, and biased anchoring to individual samples in high-dimensional multimodal spaces.  
We address robust intrinsic multimodal knowledge editing by explicitly targeting generalization. We formalize robustness through knowledge units that group semantically equivalent multimodal inputs and define generality as consistent predictions within each unit. To expose fragile semantic regions, we introduce Latent Adversarial Robustification (LAR), which generates adversarial yet semantically coherent variants in the joint latent space. We further propose Rank-Constrained Subspace Learning (RCSL), enforcing low-rank alignment of adversarial representations at the edit layer via a singular value–based objective.  
Extensive analysis demonstrates the effectiveness of ASAM empirically. 
\end{abstract}
\addtocontents{toc}{\protect\setcounter{tocdepth}{-1}}

\section{Introduction}

Multimodal large language models (MLLMs) \cite{li2023llava,bai2025qwen2} rapidly advance generative AI across applications due to their superior cross-modal comprehension.
Knowledge Editing (KE) \cite{wang2024knowledge} has gained prominence as an efficient means to update knowledge and correct errors in pre-trained models, particularly for MLLMs where editing cross-modal structures enables precise, localized model updates.
Existing editing methods for MLLMs fall into two types \cite{pan2024towards}, \textit{external knowledge resorting} \cite{mitchell2022memory,zheng2023can} and \textit{intrinsic knowledge editing} \cite{mitchell2021fast,huang2023transformer}.
The former retrieves edited information from external memories, while the latter directly modifies internal parameters.

While external knowledge resorting retrieves contextual information from auxiliary memories, intrinsic knowledge editing directly updates model parameters to internalize new knowledge, yielding more persistent and robust edits by eliminating dependence on runtime retrieval.
Existing intrinsic editing \cite{pan2024towards,guo2025balancedit} achieve \textit{high reliability} on the edited facts and \textit{strong locality} for out‑of‑scope knowledge, but they often exhibit \textbf{limited generality}, i.e., failing to propagate updates to semantically analogous inputs, particularly under varied visual or linguistic rephrasings of the original edit.
%
%
Therefore, in this paper, we focus on addressing the performance gap in intrinsic multimodal editing by preserving reliability and locality while significantly improving generality, enabling robust and semantically adaptive editing across diverse cross-modal contexts.
However, solving the observed gap is non-trivial due to the following challenges.

\nosection{CH1. Lack of explicit and scalable supervision for semantic generalization in high-dimensional multimodal spaces}
Generalization in multimodal editing cannot be reliably enforced through a simple supervised objective.
Although rephrased inputs may appear semantically equivalent at the linguistic level or in high-level cross-attention patterns, their corresponding joint vision–language representations often reside in distant regions of a highly curved manifold.
Existing methods \cite{wang2024wise,fang2024alphaedit} perform a localized deformation of model parameters around limited editing samples without generalization constraints, which induce only a narrow semantic influence region, rendering the edited knowledge brittle to distributional variations induced by changes in visual composition, background context, or attribute phrasing.
%

\nosection{CH2. Binary editing scope assumptions hinder adaptive boundary learning under multimodal distribution shifts}
Most intrinsic editing \cite{mitchell2021fast,pan2024towards} partition samples pair-wise into \textit{in-scope} and \textit{out-of-scope} sets, implicitly assuming a static boundary.
This binary formulation oversimplifies the inherently continuous nature of semantic relatedness in multimodal inputs.
As a result, MLLMs treat editing scope as a hard decision rather than a learnable and adaptive boundary.
Moreover, intrinsic parameter updates tend to amplify the model’s reliance on shortcut correlations.
Since editing modifies entrenched correlation pathways instead of disentangled semantic factors, the model struggles to correctly extrapolate when rephrased images disrupt these correlations.
This leads to erroneous boundary judgments and exacerbates generalization failures.

\nosection{CH3. Unreliable semantic anchoring under biased editing samples}
For a single editing intent, multimodal samples often exhibit biased or incomplete distributions in the semantic space, making them unreliable references for generalization.
Previous methods \cite{guo2025balancedit} anchor parameter updates to individual samples, causing the edited knowledge to absorb incidental visual patterns or linguistic artifacts.
Such sample-centric anchoring encourages overfitting to spurious attributes and restricts the generalization boundary.
A mechanism is needed to abstract beyond imperfect sample realizations and capture the shared semantic space underlying diverse perturbations.
Otherwise, editing is confined to the biased support of observed samples, limiting robustness to rephrased inputs.

%

To tackle these challenges, we propose an \underline{A}dversarial \underline{S}ubspace \underline{A}lign\underline{M}ent based framework (\modelname) for robust multimodal knowledge editing.
The core insight of \modelname~is to move beyond sample-centric parameter updates and instead internalize edited knowledge as a consistent semantic concept that remains stable across semantically equivalent multimodal variations.
\modelname~consists of three tightly coupled components that operate at the representation level.
To address the lack of scalable semantic supervision (\textbf{{CH1}}), we introduce Latent Adversarial Robustification (LAR), a model-internal, gradient-driven mechanism that generates diverse yet semantically coherent variants in the joint vision–language latent space. By expanding the semantic neighborhood around the edited concept, LAR increases the diversity of logically equivalent edit paths without relying on external supervision.
To avoid unreliable sample anchoring (\textbf{{CH3}}), we propose Rank-Constrained Subspace Learning (RCSL), which enforces low-rank alignment across adversarial variants instead of treating any individual sample as an absolute positive. This encourages the model to implicitly discover a shared semantic subspace that captures the common semantic core of the edited knowledge under its maximal generalization boundary.
To mitigate rigid binary editing scopes (\textbf{{CH2}}), we further introduce an Asymmetric Gradient Flow that provides weak, self-supervised signals at the multi-variant level, enabling adaptive boundary shaping along meaningful semantic directions rather than enforcing instance-specific edits.
Together, these components embed edited knowledge into a low-rank semantic structure that generalizes across distribution shifts without sacrificing reliability or locality.

Our contributions can be summarized as:
(1) We propose an adversarial subspace alignment–based framework for robust multimodal knowledge editing.
(2) We introduce latent adversarial robustification to generate diverse and coherent latent variants, enabling scalable semantic supervision for editing.
(3) We develop rank-constrained subspace learning with asymmetric gradient flow, which enforces low-rank semantic alignment and adaptive boundary shaping to overcome rigid editing scopes and unreliable sample anchoring.
(4) Extensive experiments on benchmark datasets demonstrate \modelname's superiority over existing methods.
\section{Related Work}

\nosection{Knowledge Editing}
Knowledge Editing (KE) modifies factual knowledge in foundation models while preserving unrelated semantics. Methods fall into \textit{external knowledge resorting} \cite{bi2024decoding,cohen2023evaluatingrippleeffectsknowledge,zheng2023can,gu2024pokemqaprogrammableknowledgeediting,mitchell2022memory,hartvigsen2023aginggracelifelongmodel,wang2024wise}, which rely on retrieval and runtime augmentation, and \textit{intrinsic editing} \cite{zhang2024instructedit,zhang2024knowledge,jiang2025anyedit,de2021editing,mitchell2021fast,hase-etal-2023-methods,yu2023meloenhancingmodelediting,tan2024massiveeditinglargelanguage,meng2022locating,meng2022mass,dong-etal-2022-calibrating,li2024pmetprecisemodelediting,ma2024untyingreversalcursebidirectional,hernandez2024inspectingeditingknowledgerepresentations,Tamayo_2024,fang2024alphaedit,deng2025editableextendknowledgeediting,pan2025preciselocalizationmemoriesfinegrained}, which modify model parameters via hypernetworks or locate-then-edit strategies for persistent, localized updates. Multimodal extensions (UniKE \cite{pan2024towards}, BalancEdit \cite{guo2025balancedit}, ODEdit \cite{su2026out}) address cross-modal entanglement but generalization remains limited.

\nosection{Multimodal Language Models}
Multimodal LLMs (MLLMs) \cite{zhu2023minigpt,li2023blip,lin2024moe,dai2023instructblip} unify visual-linguistic representations, enabling perception, reasoning, and generation. Efficient KE is critical to correct outdated or biased knowledge \cite{li2024mike,zeng2024visual}, yet entangled representations cause edits to overfit instance-specific inputs, limiting generalization \cite{cheng2023can,huang2024vlkeb,zhang2024mc,du2025mmke}.

Detailed discussions of related work are presented in Appendix~\ref{app:related_work}. 

\section{Problem Formulation}
\label{sec:setting}
%
\nosection{Multi-modal Knowledge Editing}
Let $f_\theta : \mathcal{X}_v \times \mathcal{X}_t \rightarrow \mathcal{Y}$ denote a pre-trained multi-modal large language model parameterized by $\theta$, where $\mathcal{X}_v$ and $\mathcal{X}_t$ represent the visual and textual input spaces, respectively. The goal of Multi-modal Knowledge Editing is to modify $f_\theta$ such that new factual knowledge is incorporated, while the model’s behavior on unrelated inputs remains invariant. Formally, we have the definition of the knowledge unit as follows:
\begin{definition}[Knowledge Unit]
A \emph{knowledge unit} $\mathcal{K}_i$ is defined as a set of semantically equivalent input pairs
$\mathcal{K}_i = \{(x_v^{(j)}, x_t^{(j)})\}_{j=1}^{m_i}$,
all of which correspond to a unique ground-truth target $y_i^* \in \mathcal{Y}$.
\end{definition}
Therefore, the knowledge base is defined as $\mathcal{K} = { \mathcal{K}_1, \mathcal{K}_2, \dots, \mathcal{K}_n }$, where each knowledge unit $\mathcal{K}_i$ corresponds to a distinct concept and consists of multiple semantically equivalent multi-modal inputs $(x_v, x_t)$ mapped to the same target output $y^*_i$.

%
%
\nosection{Robust Knowledge Editing}
Conventional knowledge editing aims to obtain updated parameters $\theta'$ with an edit request $(x_v, x_t, y^*) \in \mathcal{K}_i$, such that
\begin{equation}
    f_{\theta'}(x_v, x_t) = y^*.
\end{equation}
However, robust knowledge editing requires that the update generalizes beyond a single input sample.

\begin{definition}[Robust Knowledge Editing]\label{sec:robust_def}
An edit to $\mathcal{K}_i$ is \emph{robust} if the updated model $f_{\theta'}$ satisfies
\begin{equation}
    f_{\theta'}(x_v, x_t) = y_i^*, \quad \forall (x_v, x_t) \in \mathcal{K}_i,
\end{equation}
while preserving the original predictions on $\mathcal{K}\setminus \mathcal{K}_i$.
\end{definition}
We define data pairs $(x_v, x_t) \in \mathcal{K}_i$ as \textit{in-scope knowledge} for knowledge unit $\mathcal{K}_i$, while pairs $(x_v, x_t) \notin \mathcal{K}_i$ as \textit{out-of-scope knowledge} for knowledge unit $\mathcal{K}_i$. 
We evaluate the editing performance using three primary metrics according to the empirical evaluation method \cite{zhang2024comprehensivestudyknowledgeediting}: 
\begin{equation}
    \left\{\begin{aligned}
    &\text{Rel.} =  \sum\nolimits_{(x_v, x_t, y^*)}\mathbb{I} \{\arg\max_{y} f_{\theta'}(y\mid x_v, x_t) = y^* \}  \\
    &\text{Loc.} = \mathbb{E}_{(x_v, x_t) \sim \mathcal{K} \setminus \mathcal{K}_i} \left[ \mathbb{I} \{ f_{\theta'}(y \mid x_v, x_t) = f_{\theta}(y \mid x_v, x_t) \} \right] \\
    &\text{Gen.} = \mathbb{E}_{(x_v, x_t, y^*) \sim \mathcal{K}_i} \left[ \mathbb{I} \{ f_{\theta'}(x_v, x_t) = y^* \} \right]
    \end{aligned}
    \right.
\end{equation}
Rel. evaluates the edit success on the specific query, Loc. verifies the preservation of unrelated knowledge, and generality Gen. measures generalization to equivalent instances. 

\begin{figure*}[t]
\centering
\includegraphics[width=0.85\linewidth]{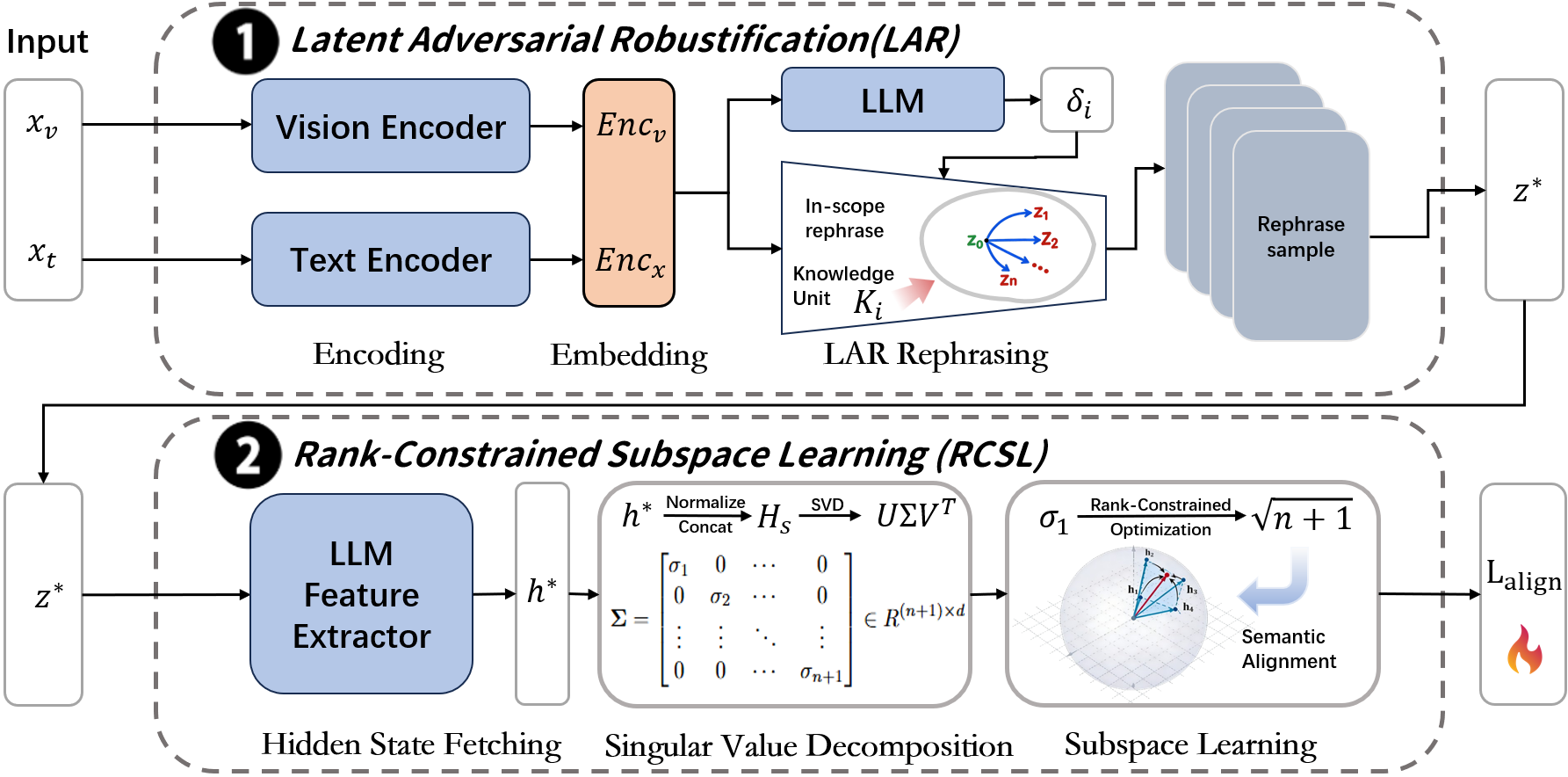}
\caption{The overall framework of \modelname, which consists of two key modules.
\ding{182} \textbf{LAR.} Given multimodal inputs, LAR perturbs input embeddings along LLM-guided gradients to generate semantically consistent rephrases. \ding{183} \textbf{RCSL.} Using these rephrases, RCSL applies SVD-based subspace learning to align editing-layer outputs, enforcing semantic consistency across variants.}
\label{story}
\vspace{-0.2 in}
\end{figure*}

\section{Methodology}
In this section, we introduce Latent Adversarial Robustification (LAR) to generate semantic variants (Sec. \ref{sec:LAR}). Rank-Constrained Subspace Learning (RCSL) is proposed to enforce representation-level consistency across these variants (Sec. \ref{sec:RCSL}). We then clarify the optimization objective with Asymmetric Gradient Flow proposed to mitigate rigid binary editing scopes (Sec. \ref{sec:optimization}). 

%
\subsection{Latent Adversarial Robustification (LAR)}\label{sec:LAR}
To evaluate and enforce the generality risk $\mathcal{R}_{\text{gen}}$, each knowledge unit $\mathcal{K}_i$ requires diverse semantic variants. However, conventional augmentation or manual paraphrasing is costly and limited to human priors, failing to expose model-specific sensitivities or \textit{blind spots}. To address this, we propose \textit{Latent Adversarial Robustification (LAR)}, a gradient-driven, model-internal method that constructs adversarial semantic neighborhoods for robust knowledge editing.
%


\nosection{Adversarial Variant Generation}
The core idea of LAR is to systematically perturb the input embeddings along directions that maximally challenge the current model parameters, thereby simulating extreme yet semantically coherent variations within the knowledge unit. Let $e_v = \text{Enc}_v(x_v)$, $e_t = \text{Enc}_t(x_t)$ denote the latent embeddings of the visual and textual components of an edit request, respectively. We concatenate them into a joint representation $z = [e_v, e_t]$.
To generate challenging variants for $\mathcal{K}_i$, we first compute the cross-entropy loss $\mathcal{L}_{ce}(f_\theta(z), y^*)$ via a forward pass. Unlike standard training, LAR maximizes this loss to induce adversarial perturbations. Following \cite{fu2025questiondifferentwordslatent}, we compute the gradient with respect to the latent input:
\begin{equation}
    g_z = \nabla_z \mathcal{L}_{ce}(f_\theta(z), y^*).
\end{equation}
\begin{definition}[Adversarial Latent Variant]
An adversarial variant $z_i^*$ is obtained by solving$z_i^* = z + \delta_i^*, \quad 
    \delta_i^* = \arg\max_{\|\delta\|_p \le \epsilon}
    \mathcal{L}_{\mathrm{ce}}(f_\theta(z+\delta),y^*)$,
where $\epsilon$ constrains semantic deviation.
\end{definition}
Guided by this gradient, we generate $n$ adversarial variants $z_i^* = [e_v + \delta_{v,i}, e_t + \delta_{t,i}]$ by solving the following inner maximization problem:
\begin{equation}
    \delta_i^* = \arg\max_{\|\delta_i\| \le \epsilon} \mathcal{L}_{ce}(f_\theta(z + \delta_i), y^*).
\end{equation}
In practice, $\delta_i^*$ is approximated via single-step projected gradient ascent with randomized initialization, yielding a set $\{z_1^*,\dots,z_n^*\}$. 

\textbf{Theoretical Justification for LAR.}
We justify LAR via both lower- and upper-bound perspectives. 
First, for \textit{Lower Bound} (Gradient–Knowledge Coupling), gradients link to the stored knowledge within model parameters. A first-order Taylor expansion $f(z+\delta) \approx f(z) + \nabla_z f(z)^\top \delta$ shows the gradients define semantically relevant directions, ensuring that small perturbations remain on the local knowledge manifold. 
Second, from an \textit{Upper Bound} (Controlled Deviation) view, local semantic consistency is explicitly enforced by $\|\delta\| \le \epsilon$ and the Lipschitz-like constraint $\|f(z+\delta)-f(z)\| \le L\epsilon$. $L$ is the Lipschitz constant. While LAR provides gradient-aligned local exploration, our RCSL further filters inconsistent directions by projecting variants onto a shared semantic subspace via rank constraints, ensuring perturbations are both knowledge-aware and semantically constrained.

\textbf{Empirical Boundary of Semantic Consistency.}
While strictly proving semantic invariance is non-trivial, we operationalize and bound this semantic shift through a quantifiable verification process. We define a stable semantic space for an input $x$ relative to the pre-edited model $M$ and its original output $y = M(x)$. By introducing a controlled perturbation $\delta$, we yield a perturbed output $y_{\delta} = M(x + \delta)$. 
To identify the precise boundary where the semantic meaning of $y_{\delta}$ deviates from $y$, we employ a bisection search over the magnitude $\|\delta\|$ using an independent, pre-trained semantic discriminator $D$. The maximum semantic-preserving perturbation budget $\epsilon^*$ is formally defined as:
\begin{equation}
    \epsilon^* = \sup \{ \|\delta\| : D(M(x), M(x + \delta)) \geq \tau \},
\end{equation}
where $\tau$ is a similarity threshold representing strict semantic equivalence. This provides a direct, reproducible measurement of the local generalization boundary for each edit.

\textbf{Training Objective for Robust Knowledge Editing.}
By generating variants that oppose standard training updates, LAR exposes the model to boundary cases around the edited knowledge. The resulting robust editing objective aggregates losses over all $n$ adversarial variants:
\begin{equation}
    \mathcal{L}_{\text{robust}}
    =
    \mathcal{L}_{\text{align}}
    \left(
    \{ f_{\theta'}(z_i^*) \}_{i=1}^{n}
    \right).
\end{equation}
This formulation expands a single edit point into a continuous adversarial region in latent space. By enforcing prediction consistency under worst-case perturbations, it encourages a smoother loss landscape, thereby internalizing the edited knowledge as robust, generalized representations rather than brittle instance-level mappings.

\subsection{Rank-Constrained Subspace Learning (RCSL)}\label{sec:RCSL} 
%
%
To overcome the brittleness of traditional knowledge editing in multimodal settings, we propose the \textit{Rank-Constrained Subspace Learning (RCSL)}. Its main insight lies in two parts: 1) Robust generalization requires learning a continuous, adaptive boundary rather than relying on a static and binary in/out-of-scope assumption. And 2) the edit must be anchored to an invariant semantic core that persists across input variations, not to potentially biased individual samples. RCSL operationalizes this by enforcing that all latent variants of an edit converge within a low-rank representation subspace. This ensures the model update captures the essential, shared concept, leading to consistent and reliable behavior on paraphrased inputs.

\nosection{Formulation}
Given an original edit request and its adversarial variants $\{z_0, z_1, \dots, z_n\}$ generated by LAR (with $z_0$ denoting the original sample), each input is independently forwarded through the MLLM. Let $h_i \in \mathbb{R}^d$ denote the hidden state extracted from a predefined semantic feature extractor $\phi$, which is composed of the edit layer $L$ and the early layers of the language model for input $z_i$. The edit layer is the sole target of parameter modification during the editing process, formally,
\begin{equation}
    h_i:=f_{\phi}(z_i) = f_{\phi}(z+\delta_i).
\end{equation}
Our objective is to globally align all samples symmetrically, while preserving the original computational graph for other editing objectives such as reliability and locality.

\nosection{Rank-Constrained Alignment}
To formalize the connection between semantic alignment and low-rank structure, we leverage the geometric properties of the Gram matrix. Let $\mathbf{H}_s = [\hat{\mathbf{h}}_0, \hat{\mathbf{h}}_1, \dots, \hat{\mathbf{h}}_n]^\top \in \mathbb{R}^{(n+1) \times d}$ denote the matrix of $\ell_2$-normalized hidden states, where $\hat{\mathbf{h}}_i = \mathbf{h}_i / \|\mathbf{h}_i\|_2$ for $i \in \{0, 1, \dots, n\}$. The corresponding Gram matrix $\mathbf{G} = \mathbf{H}_s \mathbf{H}_s^\top \in \mathbb{R}^{(n+1) \times (n+1)}$ captures the pairwise inner products between all normalized hidden states.

The entries in $\mathbf{G}$ can represent the cosine similarities between representations as well, thereby characterizing their geometric alignment~\cite{cicchetti2025gramianmultimodalrepresentationlearning, liu2025principledmultimodalrepresentationlearning, liu2025calibrated}. 
From Definition~\ref{sec:robust_def}, the robustness objective indicates that all adversarial variants produce identical representations:
\begin{equation}
    \hat{\mathbf{h}}_i = \hat{\mathbf{h}}_j, \quad \forall i, j \in \{0, 1, \dots, n\}.
    \label{eq:alignment_goal}
\end{equation}
Therefore, achieving this perfect alignment is equivalent to requiring $\mathbf{G} = \mathbf{1}\mathbf{1}^\top$, where $\mathbf{1} \in \mathbb{R}^{n+1}$ denotes the all-ones vector. Notably, matrix $\mathbf{G}$ has rank one, establishing a direct connection between semantic consistency and rank constraints. We formalize this observation in the following proposition:

\begin{proposition}[Rank-Constrained Alignment]
\label{prop:rank_alignment}
Let $\{\hat{\mathbf{h}}_0, \hat{\mathbf{h}}_1, \dots, \hat{\mathbf{h}}_n\}$ denote the $\ell_2$-normalized hidden states from the editing layer, and define the Gram matrix $\mathbf{G} \in \mathbb{R}^{(n+1) \times (n+1)}$ with entries $\mathbf{G}_{i,j} = \langle \hat{\mathbf{h}}_i, \hat{\mathbf{h}}_j \rangle$. Then the following statements are equivalent: 1) All hidden states are perfectly aligned, i.e., $\hat{\mathbf{h}}_i = \hat{\mathbf{h}}_j\ \forall i, j \in \{0, 1, \dots, n\}$, 2) $\mathbf{G}_{i,j} = 1\ \forall i, j \in \{0, 1, \dots, n\}$, and 3) $\mathrm{rank}(\mathbf{G}) = 1$.
\end{proposition}

\textit{Remark.} 
This rank-1 structure captures the desired invariance property: despite adversarial perturbations in the input space, the editing layer is expected to encode a shared underlying semantic concept. This naturally leads to the objective of robust knowledge editing: for any samples $(x_v, x_t)$ belonging to the same knowledge unit $\mathcal{K}$, the edited output $y^*$ should remain consistent.


\subsection{Asymmetric Gradient Flow for Optimization}
\label{sec:optimization}

To operationalize Proposition~\ref{prop:rank_alignment}, we leverage the singular value decomposition (SVD) of the hidden state matrix $\mathbf{H}_s$. Specifically, we compute the singular values $\sigma_1 \ge \sigma_2 \ge \dots \ge \sigma_{n+1}$ of $\mathbf{H}_s$ as:
$\text{SVD}(\mathbf{H}_s) = \mathbf{U} \mathbf{\Sigma} \mathbf{V}^\top, \mathbf{\Sigma} = \text{diag}(\sigma_1, \sigma_2, \dots, \sigma_{n+1})$.
Wherein, $\mathbf{U} \in \mathbb{R}^{(n+1) \times (n+1)}, \mathbf{V} \in \mathbb{R}^{d \times d}$ are orthogonal matrices. The dominance of the leading singular value $\sigma_1$ relative to the others serves as a quantitative measure of alignment quality: when $\sigma_1 \gg \sigma_j$ for $j > 1$, the representations are nearly rank-1 and thus well-aligned.
We treat the singular values as unnormalized logits and define a temperature-scaled softmax objective for subspace alignment:
\begin{equation}
    \mathcal{L}_{\text{align}} = -\log \left( \frac{\exp(\sigma_1 / \tau)}{\sum_{j=1}^{n+1} \exp(\sigma_j / \tau)} \right),
    \label{eq:align_loss}
\end{equation}
where $\tau > 0$ is a temperature hyperparameter that controls the sharpness of the alignment constraint. Minimizing $\mathcal{L}_{\text{align}}$ explicitly suppresses secondary singular directions $\{\sigma_j\}_{j>1}$, thereby encouraging all adversarial representations toward a single shared semantic axis within the editing layer.


Our method asymmetrically treats gradient flow. The original hidden state $\hat{\mathbf{h}}_0$ acts as a semantic anchor and is \emph{detached} from $\mathcal{L}_{\text{align}}$, preserving its graph for other objectives.  
All adversarial variants $\{\hat{\mathbf{h}}_i\}_{i=1}^n$ propagate gradients through $\mathcal{L}_{\text{align}}$, updating only the editing layer parameters $\theta_L$ to align diverse inputs with the original semantics:
\begin{equation}
    \nabla_{\theta_L} \mathcal{L}_{\text{align}} = \sum_{i=1}^{n} \frac{\partial \mathcal{L}_{\text{align}}}{\partial \hat{\mathbf{h}}_i} \frac{\partial \hat{\mathbf{h}}_i}{\partial \theta_L},
    \label{eq:gradient_flow}
\end{equation}
where the summation excludes $i=0$. This asymmetric gradient flow ensures that robustness is achieved through adaptive parameter updates rather than distortion of the original representation.

The overall optimization objective for robust knowledge editing is formally represented as:
\begin{equation}
    \mathcal{L}_{\text{total}} = \mathcal{L}_{\text{rel}} + \mathcal{L}_{\text{loc}} + \beta \mathcal{L}_{\text{align}},
    \label{eq:total_loss}
\end{equation}
where $\beta > 0$ balances task-specific fidelity and robustness on the representation level. Reliability and locality losses $\mathcal{L}_{\text{rel}}$ and $\mathcal{L}_{\text{loc}}$ follow standard knowledge editing formulations, ensuring accurate and localized edits. The alignment loss $\mathcal{L}_{\text{align}}$ (Eq.~\ref{eq:align_loss}) acts on hidden states from adversarial variants $\{z_i\}_{i=1}^n$ generated by LAR, aligning their representations along a shared semantic direction.


%


\nosection{Synergies}
The core robustness of \modelname~stems from the joint optimization of Latent Adversarial Robustification (LAR) and Rank-Constrained Subspace Learning (RCSL). LAR acts as an end-to-end probe, directly uncovering high-dimensional, machine-centric vulnerabilities in the latent space without the computational overhead of external LLMs or diffusion models. RCSL processes these LAR-generated variants, constraining the edited knowledge within a stable low-rank region to resolve the exposed inconsistencies.
Together, this tight coupling of adversarial variant generation and rank-constrained alignment ensures that edits generalize seamlessly across challenging multimodal inputs while preserving precise semantic consistency. Our ablation experiments further validate this synergistic effect (Sec. \ref{sec:ablation}).

\section{Experiments}
\label{sec:experiment}

\begin{table}[t]
\centering
\caption{\textbf{Performance comparison on E-VQA and E-IC tasks.} We report metrics for two MLLM backbones, i.e., BLIP2 and MiniGPT-4, across various editing methods.}

\label{tab:single_edit}
\small
\setlength{\tabcolsep}{3pt} 

\begin{tabular}{l l cccc cccc}
\toprule
& & \multicolumn{4}{c}{\textbf{E-VQA}} & \multicolumn{4}{c}{\textbf{E-IC}} \\
\cmidrule(lr){3-6} \cmidrule(lr){7-10}
\textbf{MLLM} & \textbf{Method}
& \textbf{Rel.}$\uparrow$ & \textbf{Gen.}$\uparrow$ & \textbf{T-Loc.}$\uparrow$ & \textbf{M-Loc.}$\uparrow$
& \textbf{Rel.}$\uparrow$ & \textbf{Gen.}$\uparrow$ & \textbf{T-Loc.}$\uparrow$ & \textbf{M-Loc.}$\uparrow$ \\
\midrule

\multirow{10}{*}{\textbf{BLIP2}}
& Pre-edited & 25.85 & 26.37 & 99.38 & 92.83 & 0.00 & 0.00 & 99.79 & 94.93 \\
& FT & 100.0 & 100.0 & 93.94 & 64.79 & 100.0 & 0.00 & 78.79 & 29.58 \\
& IKE & 99.71 & 99.62 & 47.74 & 2.53 & 94.40 & 88.00 & 50.43 & 2.87 \\
& SERAC & 97.60 & 97.30 & 100.0 & 3.21 & 99.71 & 99.71 & 100.0 & 2.64 \\
& T-Patcher & 80.35 & 77.82 & 87.14 & 85.28 & 72.78 & 72.75 & 71.59 & 80.49 \\
& UniKE & 94.32 & 87.18 & 95.98 & 93.15 & 74.01 & 73.84 & 76.09 & 82.36 \\
\cmidrule{2-10}
& WISE & 99.00 & 84.80 & 100.0 & 8.99 & 100.0 & 82.54 & 99.84 & 10.28 \\

& \cellcolor[rgb]{0.949,0.949,0.949}\modelname$_{\text{WISE}}$
& \cellcolor[rgb]{0.949,0.949,0.949}\textbf{99.00}
& \cellcolor[rgb]{0.949,0.949,0.949}\textbf{93.97}
& \cellcolor[rgb]{0.949,0.949,0.949}\textbf{100.0}
& \cellcolor[rgb]{0.949,0.949,0.949}\textbf{9.73}
& \cellcolor[rgb]{0.949,0.949,0.949}\textbf{100.0}
& \cellcolor[rgb]{0.949,0.949,0.949}\textbf{86.22}
& \cellcolor[rgb]{0.949,0.949,0.949}\textbf{100.0}
& \cellcolor[rgb]{0.949,0.949,0.949}\textbf{10.74} \\

\cmidrule{2-10}
& MEND & 95.22 & 94.17 & 99.53 & 92.35 & 76.10 & 60.30 & 98.18 & 79.16 \\

& \cellcolor[rgb]{0.949,0.949,0.949}\modelname$_{\text{MEND}}$
& \cellcolor[rgb]{0.949,0.949,0.949}\textbf{95.99}
& \cellcolor[rgb]{0.949,0.949,0.949}\textbf{94.65}
& \cellcolor[rgb]{0.949,0.949,0.949}\textbf{99.66}
& \cellcolor[rgb]{0.949,0.949,0.949}\textbf{93.80}
& \cellcolor[rgb]{0.949,0.949,0.949}\textbf{77.40}
& \cellcolor[rgb]{0.949,0.949,0.949}\textbf{62.60}
& \cellcolor[rgb]{0.949,0.949,0.949}\textbf{98.29}
& \cellcolor[rgb]{0.949,0.949,0.949}79.07 \\

\midrule

\multirow{10}{*}{\textbf{MiniGPT-4}}
& Pre-edited & 19.21 & 24.08 & 99.44 & 91.56 & 0.00 & 0.00 & 99.79 & 94.93 \\
& FT & 100.0 & 100.0 & 97.50 & 40.85 & 100.0 & 0.00 & 95.00 & 39.83 \\
& IKE & 99.95 & 99.90 & 50.02 & 3.31 & 90.30 & 90.00 & 51.49 & 4.27 \\
& SERAC & 91.70 & 98.60 & 99.99 & 3.72 & 83.60 & 93.10 & 99.99 & 4.65 \\
& T-Patcher & 70.56 & 68.79 & 64.45 & 81.77 & 69.54 & 68.95 & 63.59 & 81.34 \\
& UniKE & 84.32 & 81.29 & 78.45 & 85.81 & 72.18 & 70.41 & 68.53 & 84.59 \\
\cmidrule{2-10}
& WISE & 100.0 & 98.86 & 100.0 & 56.48 & 95.68 & 83.01 & 100.0 & 89.80 \\

& \cellcolor[rgb]{0.949,0.949,0.949}\modelname$_{\text{WISE}}$
& \cellcolor[rgb]{0.949,0.949,0.949}\textbf{100.0}
& \cellcolor[rgb]{0.949,0.949,0.949}\textbf{99.00}
& \cellcolor[rgb]{0.949,0.949,0.949}\textbf{100.0}
& \cellcolor[rgb]{0.949,0.949,0.949}\textbf{56.75}
& \cellcolor[rgb]{0.949,0.949,0.949}\textbf{95.84}
& \cellcolor[rgb]{0.949,0.949,0.949}\textbf{83.36}
& \cellcolor[rgb]{0.949,0.949,0.949}\textbf{100.0}
& \cellcolor[rgb]{0.949,0.949,0.949}88.49 \\

\cmidrule{2-10}
& MEND & 96.13 & 95.41 & 99.61 & 93.92 & 82.00 & 59.60 & 99.17 & 87.17 \\

& \cellcolor[rgb]{0.949,0.949,0.949}\modelname$_{\text{MEND}}$
& \cellcolor[rgb]{0.949,0.949,0.949}\textbf{96.60}
& \cellcolor[rgb]{0.949,0.949,0.949}\textbf{95.80}
& \cellcolor[rgb]{0.949,0.949,0.949}\textbf{99.62}
& \cellcolor[rgb]{0.949,0.949,0.949}93.60
& \cellcolor[rgb]{0.949,0.949,0.949}\textbf{82.80}
& \cellcolor[rgb]{0.949,0.949,0.949}\textbf{61.40}
& \cellcolor[rgb]{0.949,0.949,0.949}99.08
& \cellcolor[rgb]{0.949,0.949,0.949}\textbf{87.64} \\

\bottomrule
\end{tabular}
\vspace{-3mm}
\end{table}

\subsection{Experimental Setups}
\nosection{Benchmarks}
Following the prior research \citep{pan2024towards}, we evaluate on the MMEdit benchmark \citep{cheng2023can}, which comprises two subtasks: Editing VQA (E-VQA) and Editing Image Caption (E-IC).
In this evaluation framework, BLIP2-OPT \citep{li2023blip} and MiniGPT-4 \citep{zhu2023minigpt} serve as our foundational MLLMs.
The performance is assessed via Reliability, Generality, and Locality (T-Loc. and M-Loc.) metrics.

\nosection{Baselines \& Implementation Details}
We select 7 baselines across four types, including fine-tuning (FT), MEND \citep{mitchell2021fast}, IKE \citep{zheng2023can}, SERAC \cite{mitchell2022memory}, T-Patcher \citep{huang2023transformer}, WISE \citep{wang2024wise},  and  UniKE \citep{pan2024towards}. How we incorporate \modelname~into each method as a plug-and-play optimization framework in Appendix \ref{app:baselines} and implementation details are in Appendix \ref{app:implementde}. The overhead performance and hyperparameter settings are in Appendix \ref{app:overhead} and Appendix \ref{app:hyperparam} respectively.
%
%

\subsection{Knowledge Editing Performance}
\nosection{Single Editing Results}
Table~\ref{tab:single_edit} compares single-step editing performance, showing:
(1) \textit{\modelname~consistently enhances generality across editing paradigms while preserving reliability and locality.}
In both WISE and MEND, \modelname~yields clear generality gains, e.g., on BLIP2 E-VQA $84.80 \rightarrow 93.97$, on E-IC $82.54 \rightarrow 86.22$, without sacrificing Rel. or Loc. This indicates that \modelname~ improves generalization by restructuring edited representations rather than expanding parameter update scope, making it complementary to both memory-based and gradient-based editors.
(2) \textit{Generality gains of \modelname~are consistent across backbones and tasks.} Gains appear on all MLLMs and both tasks, despite differences in architecture, pretraining, and data. This shows \modelname~works at a largely model- and task-agnostic representation level, allowing edits to generalize across heterogeneous multimodal settings, even when the base model has low reliability, locality, or generality.

\nosection{Sequential Editing Results}
Lifelong knowledge editing is a harder scenario as errors introduced by earlier edits may accumulate and propagate to later updates.
Following~\cite{wang2024wise}, we conduct sequential editing with steps of $\{1,10,100\}$.
The results in Table~\ref{tab:sequential_edit} validates:
(1) \textit{Performance advantage under long edit sequences.}
\modelname~consistently outperforms WISE in sequential editing, with the gap widening as the number of edits increases, e.g., Gen. increase from $34.90 \rightarrow 50.90$ on E-VQA at $n=100$ while Loc. maintains. This indicates that \modelname~is more robust to semantic interference, reducing cross-edit disruption compared to WISE.
(2) \textit{Stability across metrics.} On E-IC, \modelname~maintains higher reliability and generality than WISE at $n=10$ and $n=100$, indicating it controls error accumulation. By limiting propagation of edit-induced perturbations, the model avoids catastrophic drift, sustaining performance over long editing sequences.

\begin{table}[t]
\centering
\small
\caption{Editing performance on sequential editing studies. \emph{n} denotes the sequential editing iterations.}
\label{tab:sequential_edit}

\begin{tabular}{c|l|cccc cccc}
\toprule
& & \multicolumn{4}{c}{\textbf{E-VQA}} & \multicolumn{4}{c}{\textbf{E-IC}} \\
\cmidrule(lr){3-6} \cmidrule(lr){7-10}
$n$ & \textbf{Method} 
& Rel.$\uparrow$ & Gen.$\uparrow$ & T-Loc.$\uparrow$ & M-Loc.$\uparrow$
& Rel.$\uparrow$ & Gen.$\uparrow$ & T-Loc.$\uparrow$ & M-Loc.$\uparrow$ \\
\midrule

\multirow{2}{*}{1}
 & WISE  & 99.00 & 84.80 & 100.0 & 8.99 
          & 100.0 & 82.54 & 99.84 & 10.28 \\
 & \modelname  & \textbf{99.00} & \textbf{93.97} & \textbf{100.0} & \textbf{9.73} 
                  & \textbf{100.0} & \textbf{86.22} & \textbf{100.0} & \textbf{10.74} \\

\midrule

\multirow{2}{*}{10}
 & WISE  & 64.33 & 52.33 & 96.64 & 8.48 
          & 90.88 & 68.50 & 95.46 & 10.15 \\
 & \modelname  & \textbf{75.33} & \textbf{58.33} & \textbf{98.07} & \textbf{8.49} 
                  & \textbf{91.37} & \textbf{82.78} & \textbf{98.79} & \textbf{10.39} \\

\midrule

\multirow{2}{*}{100}
 & WISE  & 48.63 & 34.90 & 88.76 & 7.33 
          & 68.50 & 56.30 & 84.10 & 9.91 \\
 & \modelname  & \textbf{51.20} & \textbf{50.90} & \textbf{99.23} & 3.38 
                  & \textbf{76.38} & \textbf{67.99} & \textbf{98.46} & 5.72 \\

\bottomrule
\end{tabular}
\vspace{-5mm}
\end{table}

\subsection{Ablation Studies}
\label{sec:ablation}
\nosection{Model Generality Results}
Table~\ref{tab:model} reports the performance of \modelname~on two representative MLLM backbones, i.e., LLaVA and Qwen2VL, from which we observe:

$\bullet$ \textit{Consistent generality improvement across different backbones without compromising reliability.}
On E-VQA, \modelname~improves generality for both models, i.e., from $64.00 \rightarrow 64.67$ on LLaVA and from $68.80 \rightarrow 69.97$ on Qwen2VL, while preserving reliability (LLaVA: $100.0 \rightarrow 100.0$, Qwen2VL: $87.15 \rightarrow 87.99$). Meanwhile, both textual and multimodal locality are consistently enhanced (e.g., T-Loc.\ $53.33 \rightarrow 54.29$ on LLaVA and $29.04 \rightarrow 30.35$ on Qwen2VL), indicating improved semantic generalization without introducing additional interference.

$\bullet$ \textit{Stable behavior under caption-based editing across models.}
On E-IC, \modelname~remains stable while improving in generality and locality. Specifically, LLaVA exhibits a slight gain in generality ($86.14 \rightarrow 86.15$) with improved locality (T-Loc.\ $41.30 \rightarrow 41.90$), while Qwen2VL shows similar trends (Gen.\ $72.27 \rightarrow 72.34$, T-Loc.\ $28.66 \rightarrow 29.62$). Reliability remains nearly unchanged for both backbones, demonstrating that \modelname~refines representations without disrupting existing knowledge.

Overall, while the absolute gains are relatively modest, the improvements across both LLaVA and Qwen2VL suggest that \modelname~is model-agnostic and generalizes effectively to diverse MLLM architectures, serving as a lightweight yet reliable enhancement for multimodal knowledge editing.

\begin{table}[t]
\centering
\small
\caption{Performance comparison of ASAM built on the LLaVA and Qwen2VL.}
\label{tab:model}

\begin{tabular}{l|cccc cccc}
\toprule
& \multicolumn{4}{c}{\textbf{E-VQA}} & \multicolumn{4}{c}{\textbf{E-IC}} \\
\cmidrule(lr){2-5} \cmidrule(lr){6-9}
\textbf{Method} 
& Rel.$\uparrow$ & Gen.$\uparrow$ & T-Loc.$\uparrow$ & M-Loc.$\uparrow$
& Rel.$\uparrow$ & Gen.$\uparrow$ & T-Loc.$\uparrow$ & M-Loc.$\uparrow$ \\
\midrule

LLaVA 
& 100.0 & 64.00 & 53.33 & 6.97  
& 99.88 & 86.14 & 41.30 & 6.22  \\

LLaVA+\modelname 
& \textbf{100.0} & \textbf{64.67} & \textbf{54.29} & \textbf{7.01} 
& \textbf{99.88} & \textbf{86.15} & \textbf{41.90} & \textbf{6.24} \\

\midrule

Qwen2VL 
& 87.15 & 68.80 & 29.04 & 7.12 
& 86.94 & 72.27 & 28.66 & 5.26 \\

Qwen2VL+\modelname
& \textbf{87.99} & \textbf{69.97} & \textbf{30.35} & \textbf{7.17}  
& \textbf{87.24} & \textbf{72.34} & \textbf{29.62} & \textbf{5.31} \\

\bottomrule
\end{tabular}
\vspace{-8mm}
\end{table}


\begin{table}[t]
\centering
\small
\caption{Ablation study on key components of \modelname.}
\label{tab:ablation}

\begin{tabular}{l|cccc cccc}
\toprule
& \multicolumn{4}{c}{\textbf{E-VQA}} & \multicolumn{4}{c}{\textbf{E-IC}} \\
\cmidrule(lr){2-5} \cmidrule(lr){6-9}
\textbf{Method} 
& Rel.$\uparrow$ & Gen.$\uparrow$ & T-Loc.$\uparrow$ & M-Loc.$\uparrow$
& Rel.$\uparrow$ & Gen.$\uparrow$ & T-Loc.$\uparrow$ & M-Loc.$\uparrow$ \\
\midrule

Original 
& 99.00 & 92.00 & 100.0 & 9.72  
& 99.61 & 84.14 & 100.0 & 10.43 \\

SimCSE 
& 99.00 & 92.00 & 100.0 & 9.29 
& 99.86 & 84.93 & 100.0 & 10.34 \\

LAR   
& \textbf{99.00} & \textbf{93.97} & \textbf{100.0} & \textbf{9.73} 
& \textbf{100.0} & \textbf{86.22} & \textbf{100.0} & \textbf{10.74} \\

\midrule

Cosine Align 
& 99.00 & 87.00 & 100.0 & 8.94 
& 100.0 & 83.05 & 99.84 & 9.95 \\

$\ell_2$-norm Align 
& 99.00 & 87.67 & 97.7 & 7.57 
& 99.87 & 83.00 & 98.57 & 8.92 \\

RCSL  
& \textbf{99.00} & \textbf{93.97} & \textbf{100.0} & \textbf{9.73} 
& \textbf{100.0} & \textbf{86.22} & \textbf{100.0} & \textbf{10.74} \\

\bottomrule
\end{tabular}
\vspace{-2mm}
\end{table}

\textbf{Effect of Adversarial Variant Generation.} 
We first investigate how different sources of semantic variants influence editing performance, where we set variants as using (a) only original samples, (b) SimCSE-based paraphrases~\cite{gao2022simcsesimplecontrastivelearning} constructed by retrieving semantically similar rephrases using sentence embedding similarity, and (c) adversarial variants generated by LAR.
From results in \ref{tab:ablation}, we observe:
(1) \textit{Limited gains from surface-level paraphrasing.}
Across both E-VQA and E-IC, SimCSE-based paraphrases yield only marginal generality improvements. This suggests that surface-level lexical or syntactic variations provide limited coverage of the underlying semantic neighborhood, leaving the model sensitive to unseen semantic directions.
(2) \textit{Generality gains from LAR.} LAR variants boost generality consistently (E-VQA $92.00\rightarrow93.97$, E-IC $84.93\rightarrow86.22$) while preserving reliability and locality, showing that LAR exposes model-sensitive semantic directions and enables edits to generalize beyond specific inputs.



\textbf{Effect of Rank-Constrained Subspace Learning.}
We further examine the impact of different representation alignment objectives applied at the edit layer, including variants as aligning with (a) cosine similarity loss, (b) $\ell_2$-norm regularization, and (c) the proposed RCSL. 
From Table \ref{tab:ablation} we can find:
(1) \textit{Limits of pairwise/magnitude alignment.} Cosine and $\ell_2$ constraints stabilize training but hurt generality and sometimes locality, showing that sample-level or magnitude-based constraints fail to capture the invariant semantic structure of adversarial variants, over-restricting representations and impeding generalization.
(2) \textit{Effectiveness of RCSL for robust knowledge internalization.}
RCSL consistently achieves the best overall performance, preserving LAR-induced generality gains while maintaining near-perfect locality across datasets. This validates the low-rank alignment hypothesis, showing that enforcing a shared principal semantic direction enables robust knowledge internalization without interfering with unrelated knowledge.

\begin{figure}[t]
    \centering
    \begin{minipage}[t]{0.45\linewidth}
        \centering
        \includegraphics[width=\textwidth]{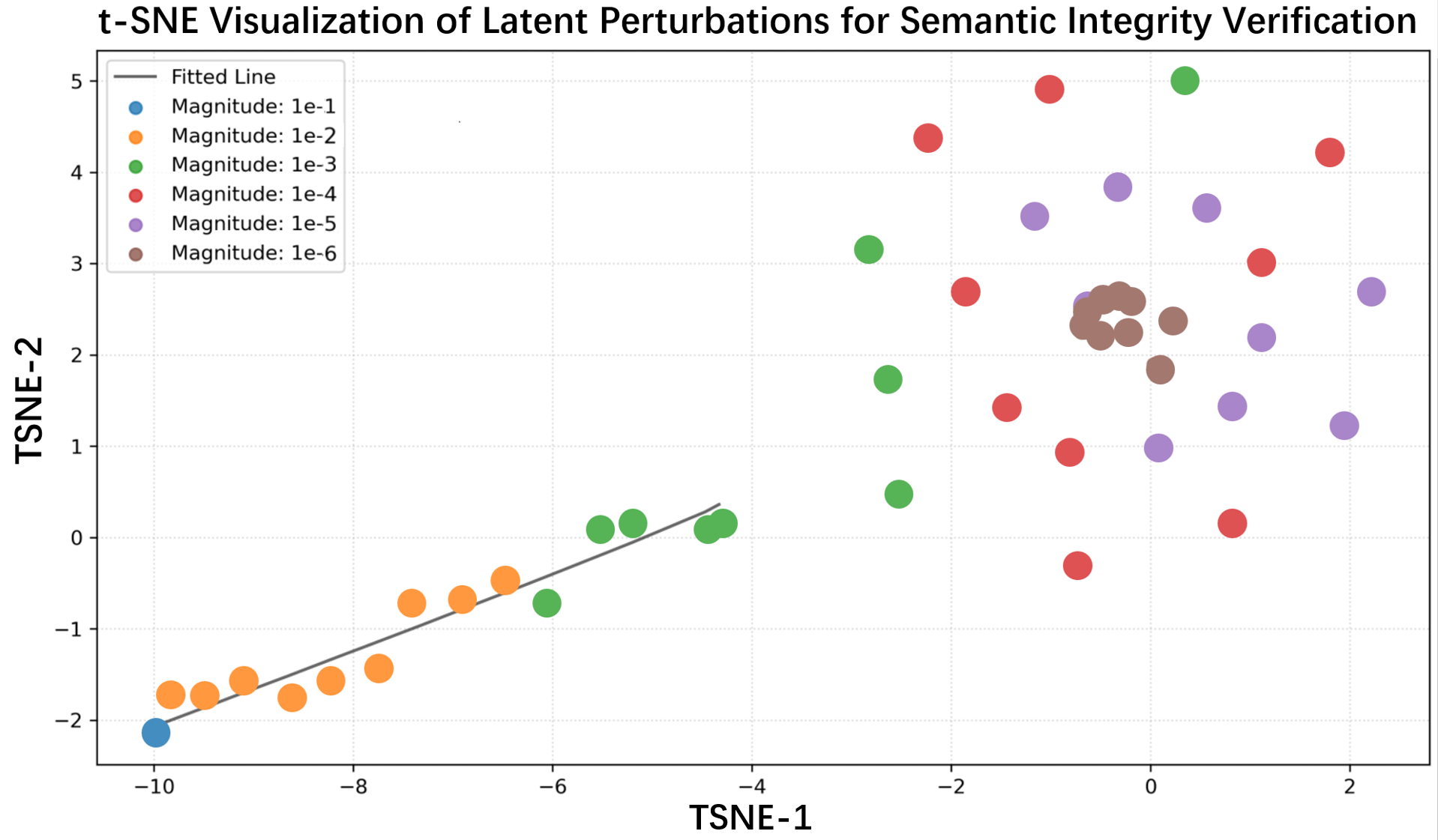}
        \vspace{-5mm}
        \caption{t-SNE visualization of representations under varying perturbations.}
        \label{fig:tsne}
    \end{minipage}
    \hfill
    \begin{minipage}[t]{0.48\linewidth}
        \centering
        \includegraphics[width=\linewidth]{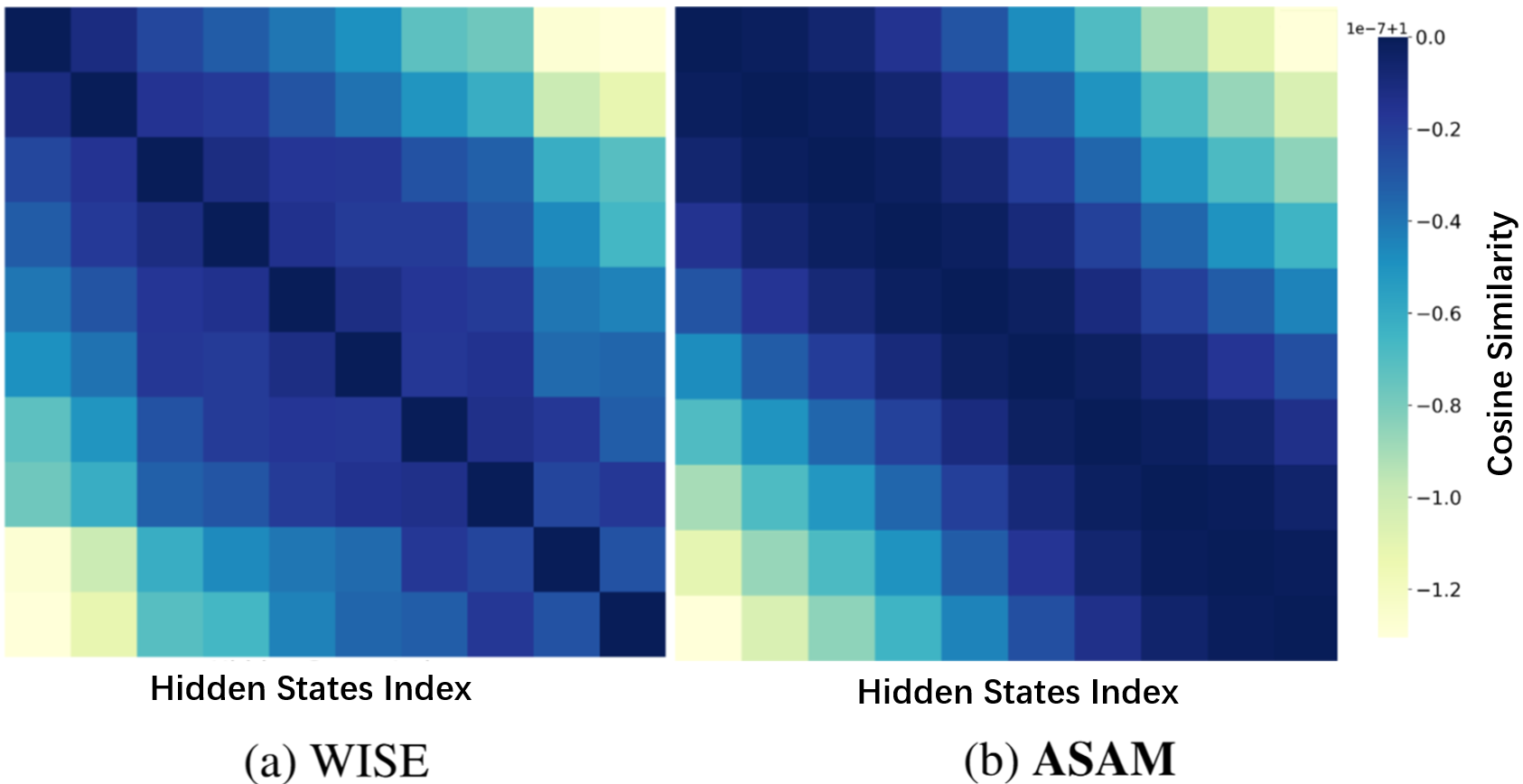}
        \caption{Heatmaps of hidden-state similarities.
        }
        \label{fig:simulation}
    \end{minipage}
    \vspace{-6mm}
\end{figure}

\nosection{Visualization of Perturbation Effect}
To illustrate effects on perturbations on model representations, we apply varying magnitudes $\epsilon$ to inputs and project the resulting latent features via t-SNE (Fig.~\ref{fig:tsne}). We observe a critical transition around $\epsilon \approx 10^{-3}$: for smaller perturbations, representations form a ring-like structure with larger $\epsilon$ samples at the periphery, indicating uniform, superficial effects that preserve semantic structure and reasoning. Beyond this threshold, representations align along a dominant direction, reflecting perturbation concentration on semantically important features, increasing model sensitivity and error likelihood. The transition defines a boundary separating ineffective perturbations from those that are semantically significant but potentially degrade model robustness.

\nosection{Visualization of Representation Consistency and Low-rank Alignment} 
\label{sec:visualization}
We visualize edit-layer representations under variations by computing pairwise cosine similarities among hidden states. 
For each sample, we generate 10 variants and present the similarity structure as a heatmap in Figure \ref{fig:simulation}, showing:
(1) \textit{Diagonal dominance from a shared principal semantic direction.}
\modelname~shows a wider band of high similarity near the diagonal, indicating semantically close inputs collapse onto a common axis via a shared low-rank direction at the edit layer.
(2) \textit{Globally elevated off-diagonal similarity under RCSL.}
The similarity matrix of \modelname~is uniformly darker beyond the diagonal, reflecting that adversarial variants are encoded within a low-rank subspace, yielding consistent representations.

\nosection{Interpretation of Generalization under Perturbations}
To assess generalization, we perturb images and texts with different scaling factors to generate variants with varying input shifts.
For each sample, we compare WISE and \modelname, evaluating if the generated descriptions correctly express the target concept (e.g., \textit{pizza}).
From Figure \ref{fig:case} in Appendix \ref{app:case}, we conclude:
(1) \textit{Generalization failure of the baseline.}
The baseline fails to recall the target in 3 of 10 cases, revealing its edited knowledge remains fragile and overly specific to the exact configuration.
(2) \textit{Robust concept preservation by \modelname.}
In contrast, \modelname~preserves the target concept across all perturbed samples. This stems from internalizing knowledge at an abstract semantic level rather than memorizing surface patterns. By encouraging invariance, it maintains stable concept activation even under substantial variations.




\section{Conclusion and Future Work}
\label{sec:conclusion}
In this paper, we presented \modelname, an adversarial subspace alignment based framework for robust multimodal knowledge editing. By combining latent adversarial robustification, \modelname~expands the semantic coverage of edits with scalable latent supervision, while rank-constrained subspace learning and asymmetric gradient flow jointly enforce low-rank alignment and adaptive boundaries, yielding improved robustness under multimodal distribution shifts. Extensive experiments demonstrate the effectiveness of \modelname. As future work, we plan to extend this framework to broader multimodal domains, such as video–language models and embodied agents, exploring its potential for robust knowledge control in more complex and dynamic environments.



\bibliographystyle{plainnat}  
\bibliography{reference}      

@article{meng2022mass,
  title={Mass-editing memory in a transformer},
  author={Meng, Kevin and Sharma, Arnab Sen and Andonian, Alex and Belinkov, Yonatan and Bau, David},
  journal={arXiv preprint arXiv:2210.07229},
  year={2022}
}

@article{li2023llava,
  title={Llava-med: Training a large language-and-vision assistant for biomedicine in one day},
  author={Li, Chunyuan and Wong, Cliff and Zhang, Sheng and Usuyama, Naoto and Liu, Haotian and Yang, Jianwei and Naumann, Tristan and Poon, Hoifung and Gao, Jianfeng},
  journal={Advances in Neural Information Processing Systems},
  volume={36},
  pages={28541--28564},
  year={2023}
}

@article{bai2025qwen2,
  title={Qwen2. 5-vl technical report},
  author={Bai, Shuai and Chen, Keqin and Liu, Xuejing and Wang, Jialin and Ge, Wenbin and Song, Sibo and Dang, Kai and Wang, Peng and Wang, Shijie and Tang, Jun and others},
  journal={arXiv preprint arXiv:2502.13923},
  year={2025}
}

@article{wang2024knowledge,
  title={Knowledge editing for large language models: A survey},
  author={Wang, Song and Zhu, Yaochen and Liu, Haochen and Zheng, Zaiyi and Chen, Chen and Li, Jundong},
  journal={ACM Computing Surveys},
  volume={57},
  number={3},
  pages={1--37},
  year={2024},
  publisher={ACM New York, NY}
}

@article{fang2024alphaedit,
  title={Alphaedit: Null-space constrained knowledge editing for language models},
  author={Fang, Junfeng and Jiang, Houcheng and Wang, Kun and Ma, Yunshan and Jie, Shi and Wang, Xiang and He, Xiangnan and Chua, Tat-Seng},
  journal={arXiv preprint arXiv:2410.02355},
  year={2024}
}

@article{jiang2025anyedit,
  title={Anyedit: Edit any knowledge encoded in language models},
  author={Jiang, Houcheng and Fang, Junfeng and Zhang, Ningyu and Ma, Guojun and Wan, Mingyang and Wang, Xiang and He, Xiangnan and Chua, Tat-seng},
  journal={arXiv preprint arXiv:2502.05628},
  year={2025}
}

@article{huang2023transformer,
  title={Transformer-patcher: One mistake worth one neuron},
  author={Huang, Zeyu and Shen, Yikang and Zhang, Xiaofeng and Zhou, Jie and Rong, Wenge and Xiong, Zhang},
  journal={arXiv preprint arXiv:2301.09785},
  year={2023}
}

@article{dai2023instructblip,
  title={Instructblip: Towards general-purpose vision-language models with instruction tuning},
  author={Dai, Wenliang and Li, Junnan and Li, Dongxu and Tiong, Anthony and Zhao, Junqi and Wang, Weisheng and Li, Boyang and Fung, Pascale N and Hoi, Steven},
  journal={Advances in neural information processing systems},
  volume={36},
  pages={49250--49267},
  year={2023}
}

@article{lin2024moe,
  title={Moe-llava: Mixture of experts for large vision-language models},
  author={Lin, Bin and Tang, Zhenyu and Ye, Yang and Huang, Jinfa and Zhang, Junwu and Pang, Yatian and Jin, Peng and Ning, Munan and Luo, Jiebo and Yuan, Li},
  journal={arXiv preprint arXiv:2401.15947},
  year={2024}
}

@article{de2021editing,
  title={Editing factual knowledge in language models},
  author={De Cao, Nicola and Aziz, Wilker and Titov, Ivan},
  journal={arXiv preprint arXiv:2104.08164},
  year={2021}
}

@misc{su2026out,
      title={Out-of-Distribution Generalization via Invariant Trajectories for Multimodal Large Language Model Editing}, 
      author={Jiajie Su and Haoyuan Wang and Xiaohua Feng and Yunshan Ma and Xiaobo Xia and Yuyuan Li and Xiaolin Zheng and Jianmao Xiao and Chaochao Chen},
      year={2026},
      eprint={2601.19700},
      archivePrefix={arXiv},
      primaryClass={cs.LG},
      url={https://arxiv.org/abs/2601.19700}, 
}

@article{cheng2023can,
  title={Can we edit multimodal large language models?},
  author={Cheng, Siyuan and Tian, Bozhong and Liu, Qingbin and Chen, Xi and Wang, Yongheng and Chen, Huajun and Zhang, Ningyu},
  journal={arXiv preprint arXiv:2310.08475},
  year={2023}
}

@article{du2025mmke,
  title={Mmke-bench: A multimodal editing benchmark for diverse visual knowledge},
  author={Du, Yuntao and Jiang, Kailin and Gao, Zhi and Shi, Chenrui and Zheng, Zilong and Qi, Siyuan and Li, Qing},
  journal={arXiv preprint arXiv:2502.19870},
  year={2025}
}

@article{pan2024towards,
  title={Towards unified multimodal editing with enhanced knowledge collaboration},
  author={Pan, Kaihang and Fan, Zhaoyu and Li, Juncheng and Yu, Qifan and Fei, Hao and Tang, Siliang and Hong, Richang and Zhang, Hanwang and Sun, Qianru},
  journal={Advances in Neural Information Processing Systems},
  volume={37},
  pages={110290--110314},
  year={2024}
}

@inproceedings{guo2025balancedit,
      title={BalancEdit: Dynamically Balancing the Generality-Locality Trade-off in Multi-modal Model Editing}, 
      author={Dongliang Guo and Mengxuan Hu and Zihan Guan and Thomas Hartvigsen and Sheng Li},
      booktitle={International Conference on Machine Learning},
      year={2025},
      url={https://arxiv.org/abs/2505.01343},
}

@article{mitchell2021fast,
  title={Fast model editing at scale},
  author={Mitchell, Eric and Lin, Charles and Bosselut, Antoine and Finn, Chelsea and Manning, Christopher D},
  journal={arXiv preprint arXiv:2110.11309},
  year={2021}
}

@article{wang2024wise,
  title={Wise: Rethinking the knowledge memory for lifelong model editing of large language models},
  author={Wang, Peng and Li, Zexi and Zhang, Ningyu and Xu, Ziwen and Yao, Yunzhi and Jiang, Yong and Xie, Pengjun and Huang, Fei and Chen, Huajun},
  journal={Advances in Neural Information Processing Systems},
  volume={37},
  pages={53764--53797},
  year={2024}
}

@inproceedings{mitchell2022memory,
  title={Memory-based model editing at scale},
  author={Mitchell, Eric and Lin, Charles and Bosselut, Antoine and Manning, Christopher D and Finn, Chelsea},
  booktitle={International Conference on Machine Learning},
  pages={15817--15831},
  year={2022},
  organization={PMLR}
}

@article{zeng2024visual,
  title={Visual-oriented fine-grained knowledge editing for multimodal large language models},
  author={Zeng, Zhen and Gu, Leijiang and Yang, Xun and Duan, Zhangling and Shi, Zenglin and Wang, Meng},
  journal={arXiv preprint arXiv:2411.12790},
  year={2024}
}

@article{zhang2024instructedit,
  title={Instructedit: Instruction-based knowledge editing for large language models},
  author={Zhang, Ningyu and Tian, Bozhong and Cheng, Siyuan and Liang, Xiaozhuan and Hu, Yi and Xue, Kouying and Gou, Yanjie and Chen, Xi and Chen, Huajun},
  journal={arXiv preprint arXiv:2402.16123},
  year={2024}
}

@article{meng2022locating,
  title={Locating and editing factual associations in gpt},
  author={Meng, Kevin and Bau, David and Andonian, Alex and Belinkov, Yonatan},
  journal={Advances in neural information processing systems},
  volume={35},
  pages={17359--17372},
  year={2022}
}

@article{zhang2024knowledge,
  title={Knowledge graph enhanced large language model editing},
  author={Zhang, Mengqi and Ye, Xiaotian and Liu, Qiang and Ren, Pengjie and Wu, Shu and Chen, Zhumin},
  journal={arXiv preprint arXiv:2402.13593},
  year={2024}
}

@article{bi2024decoding,
  title={Decoding by Contrasting Knowledge: Enhancing LLMs' Confidence on Edited Facts},
  author={Bi, Baolong and Liu, Shenghua and Mei, Lingrui and Wang, Yiwei and Ji, Pengliang and Cheng, Xueqi},
  journal={arXiv preprint arXiv:2405.11613},
  year={2024}
}

@article{zheng2023can,
  title={Can we edit factual knowledge by in-context learning?},
  author={Zheng, Ce and Li, Lei and Dong, Qingxiu and Fan, Yuxuan and Wu, Zhiyong and Xu, Jingjing and Chang, Baobao},
  journal={arXiv preprint arXiv:2305.12740},
  year={2023}
}

@inproceedings{li2023blip,
  title={Blip-2: Bootstrapping language-image pre-training with frozen image encoders and large language models},
  author={Li, Junnan and Li, Dongxu and Savarese, Silvio and Hoi, Steven},
  booktitle={International conference on machine learning},
  pages={19730--19742},
  year={2023},
  organization={PMLR}
}

@article{huang2024vlkeb,
  title={Vlkeb: A large vision-language model knowledge editing benchmark},
  author={Huang, Han and Zhong, Haitian and Yu, Tao and Liu, Qiang and Wu, Shu and Wang, Liang and Tan, Tieniu},
  journal={Advances in Neural Information Processing Systems},
  volume={37},
  pages={9257--9280},
  year={2024}
}

@article{li2024mike,
  title={Mike: A new benchmark for fine-grained multimodal entity knowledge editing},
  author={Li, Jiaqi and Du, Miaozeng and Zhang, Chuanyi and Chen, Yongrui and Hu, Nan and Qi, Guilin and Jiang, Haiyun and Cheng, Siyuan and Tian, Bozhong},
  journal={arXiv preprint arXiv:2402.14835},
  year={2024}
}

@article{zhang2024mc,
  title={Mc-mke: A fine-grained multimodal knowledge editing benchmark emphasizing modality consistency},
  author={Zhang, Junzhe and Zhang, Huixuan and Yin, Xunjian and Huang, Baizhou and Zhang, Xu and Hu, Xinyu and Wan, Xiaojun},
  journal={arXiv preprint arXiv:2406.13219},
  year={2024}
}

@article{zhu2023minigpt,
  title={Minigpt-4: Enhancing vision-language understanding with advanced large language models},
  author={Zhu, Deyao and Chen, Jun and Shen, Xiaoqian and Li, Xiang and Elhoseiny, Mohamed},
  journal={arXiv preprint arXiv:2304.10592},
  year={2023}
}

@misc{fu2025questiondifferentwordslatent,
      title={Same Question, Different Words: A Latent Adversarial Framework for Prompt Robustness}, 
      author={Tingchen Fu and Fazl Barez},
      year={2025},
      eprint={2503.01345},
      archivePrefix={arXiv},
      primaryClass={cs.CL},
      url={https://arxiv.org/abs/2503.01345}, 
}

@misc{zhang2024comprehensivestudyknowledgeediting,
      title={A Comprehensive Study of Knowledge Editing for Large Language Models}, 
      author={Ningyu Zhang and Yunzhi Yao and Bozhong Tian and Peng Wang and Shumin Deng and Mengru Wang and Zekun Xi and Shengyu Mao and Jintian Zhang and Yuansheng Ni and Siyuan Cheng and Ziwen Xu and Xin Xu and Jia-Chen Gu and Yong Jiang and Pengjun Xie and Fei Huang and Lei Liang and Zhiqiang Zhang and Xiaowei Zhu and Jun Zhou and Huajun Chen},
      year={2024},
      eprint={2401.01286},
      archivePrefix={arXiv},
      primaryClass={cs.CL},
      url={https://arxiv.org/abs/2401.01286}, 
}

@misc{gao2022simcsesimplecontrastivelearning,
      title={SimCSE: Simple Contrastive Learning of Sentence Embeddings}, 
      author={Tianyu Gao and Xingcheng Yao and Danqi Chen},
      year={2022},
      eprint={2104.08821},
      archivePrefix={arXiv},
      primaryClass={cs.CL},
      url={https://arxiv.org/abs/2104.08821}, 
}

@misc{cicchetti2025gramianmultimodalrepresentationlearning,
      title={Gramian Multimodal Representation Learning and Alignment}, 
      author={Giordano Cicchetti and Eleonora Grassucci and Luigi Sigillo and Danilo Comminiello},
      year={2025},
      eprint={2412.11959},
      archivePrefix={arXiv},
      primaryClass={cs.CV},
      url={https://arxiv.org/abs/2412.11959}, 
}

@misc{liu2025principledmultimodalrepresentationlearning,
  title={Principled multimodal representation learning},
  author={Liu, Xiaohao and Xia, Xiaobo and Ng, See-Kiong and Chua, Tat-Seng},
  journal={IEEE Transactions on Pattern Analysis and Machine Intelligence},
  year={2026},
  publisher={IEEE}
}

@article{liu2025calibrated,
  title={Calibrated Multimodal Representation Learning with Missing Modalities},
  author={Liu, Xiaohao and Xia, Xiaobo and Wei, Jiaheng and Yang, Shuo and Su, Xiu and Ng, See-Kiong and Chua, Tat-Seng},
  journal={arXiv preprint arXiv:2511.12034},
  year={2025}
}

@inproceedings{Tamayo_2024,
   title={Mass-Editing Memory with Attention in Transformers: A cross-lingual exploration of knowledge},
   url={http://dx.doi.org/10.18653/v1/2024.findings-acl.347},
   DOI={10.18653/v1/2024.findings-acl.347},
   booktitle={Findings of the Association for Computational Linguistics ACL 2024},
   publisher={Association for Computational Linguistics},
   author={Tamayo, Daniel and Gonzalez-Agirre, Aitor and Hernando, Javier and Villegas, Marta},
   year={2024},
   pages={5831–5847} }

@misc{pan2025preciselocalizationmemoriesfinegrained,
      title={Precise Localization of Memories: A Fine-grained Neuron-level Knowledge Editing Technique for LLMs}, 
      author={Haowen Pan and Xiaozhi Wang and Yixin Cao and Zenglin Shi and Xun Yang and Juanzi Li and Meng Wang},
      year={2025},
      eprint={2503.01090},
      archivePrefix={arXiv},
      primaryClass={cs.CL},
      url={https://arxiv.org/abs/2503.01090}, 
}

@misc{deng2025editableextendknowledgeediting,
      title={Everything is Editable: Extend Knowledge Editing to Unstructured Data in Large Language Models}, 
      author={Jingcheng Deng and Zihao Wei and Liang Pang and Hanxing Ding and Huawei Shen and Xueqi Cheng},
      year={2025},
      eprint={2405.15349},
      archivePrefix={arXiv},
      primaryClass={cs.CL},
      url={https://arxiv.org/abs/2405.15349}, 
}

@misc{cohen2023evaluatingrippleeffectsknowledge,
      title={Evaluating the Ripple Effects of Knowledge Editing in Language Models}, 
      author={Roi Cohen and Eden Biran and Ori Yoran and Amir Globerson and Mor Geva},
      year={2023},
      eprint={2307.12976},
      archivePrefix={arXiv},
      primaryClass={cs.CL},
      url={https://arxiv.org/abs/2307.12976}, 
}

@misc{gu2024pokemqaprogrammableknowledgeediting,
      title={PokeMQA: Programmable knowledge editing for Multi-hop Question Answering}, 
      author={Hengrui Gu and Kaixiong Zhou and Xiaotian Han and Ninghao Liu and Ruobing Wang and Xin Wang},
      year={2024},
      eprint={2312.15194},
      archivePrefix={arXiv},
      primaryClass={cs.CL},
      url={https://arxiv.org/abs/2312.15194}, 
}

@inproceedings{dong-etal-2022-calibrating,
    title = "Calibrating Factual Knowledge in Pretrained Language Models",
    author = "Dong, Qingxiu  and
      Dai, Damai  and
      Song, Yifan  and
      Xu, Jingjing  and
      Sui, Zhifang  and
      Li, Lei",
    editor = "Goldberg, Yoav  and
      Kozareva, Zornitsa  and
      Zhang, Yue",
    booktitle = "Findings of the Association for Computational Linguistics: EMNLP 2022",
    month = dec,
    year = "2022",
    address = "Abu Dhabi, United Arab Emirates",
    publisher = "Association for Computational Linguistics",
    url = "https://aclanthology.org/2022.findings-emnlp.438/",
    doi = "10.18653/v1/2022.findings-emnlp.438",
    pages = "5937--5947"
}

@misc{yu2023meloenhancingmodelediting,
      title={MELO: Enhancing Model Editing with Neuron-Indexed Dynamic LoRA}, 
      author={Lang Yu and Qin Chen and Jie Zhou and Liang He},
      year={2023},
      eprint={2312.11795},
      archivePrefix={arXiv},
      primaryClass={cs.CL},
      url={https://arxiv.org/abs/2312.11795}, 
}

@misc{hernandez2024inspectingeditingknowledgerepresentations,
      title={Inspecting and Editing Knowledge Representations in Language Models}, 
      author={Evan Hernandez and Belinda Z. Li and Jacob Andreas},
      year={2024},
      eprint={2304.00740},
      archivePrefix={arXiv},
      primaryClass={cs.CL},
      url={https://arxiv.org/abs/2304.00740}, 
}

@misc{hartvigsen2023aginggracelifelongmodel,
      title={Aging with GRACE: Lifelong Model Editing with Discrete Key-Value Adaptors}, 
      author={Thomas Hartvigsen and Swami Sankaranarayanan and Hamid Palangi and Yoon Kim and Marzyeh Ghassemi},
      year={2023},
      eprint={2211.11031},
      archivePrefix={arXiv},
      primaryClass={cs.LG},
      url={https://arxiv.org/abs/2211.11031}, 
}

@inproceedings{hase-etal-2023-methods,
    title = "Methods for Measuring, Updating, and Visualizing Factual Beliefs in Language Models",
    author = "Hase, Peter  and
      Diab, Mona  and
      Celikyilmaz, Asli  and
      Li, Xian  and
      Kozareva, Zornitsa  and
      Stoyanov, Veselin  and
      Bansal, Mohit  and
      Iyer, Srinivasan",
    editor = "Vlachos, Andreas  and
      Augenstein, Isabelle",
    booktitle = "Proceedings of the 17th Conference of the European Chapter of the Association for Computational Linguistics",
    month = may,
    year = "2023",
    address = "Dubrovnik, Croatia",
    publisher = "Association for Computational Linguistics",
    url = "https://aclanthology.org/2023.eacl-main.199/",
    doi = "10.18653/v1/2023.eacl-main.199",
    pages = "2714--2731"
}

@misc{tan2024massiveeditinglargelanguage,
      title={Massive Editing for Large Language Models via Meta Learning}, 
      author={Chenmien Tan and Ge Zhang and Jie Fu},
      year={2024},
      eprint={2311.04661},
      archivePrefix={arXiv},
      primaryClass={cs.CL},
      url={https://arxiv.org/abs/2311.04661}, 
}

@misc{li2024pmetprecisemodelediting,
      title={PMET: Precise Model Editing in a Transformer}, 
      author={Xiaopeng Li and Shasha Li and Shezheng Song and Jing Yang and Jun Ma and Jie Yu},
      year={2024},
      eprint={2308.08742},
      archivePrefix={arXiv},
      primaryClass={cs.CL},
      url={https://arxiv.org/abs/2308.08742}, 
}

@misc{ma2024untyingreversalcursebidirectional,
      title={Untying the Reversal Curse via Bidirectional Language Model Editing}, 
      author={Jun-Yu Ma and Jia-Chen Gu and Zhen-Hua Ling and Quan Liu and Cong Liu},
      year={2024},
      eprint={2310.10322},
      archivePrefix={arXiv},
      primaryClass={cs.CL},
      url={https://arxiv.org/abs/2310.10322}, 
}

@misc{radford2021learningtransferablevisualmodels,
      title={Learning Transferable Visual Models From Natural Language Supervision}, 
      author={Alec Radford and Jong Wook Kim and Chris Hallacy and Aditya Ramesh and Gabriel Goh and Sandhini Agarwal and Girish Sastry and Amanda Askell and Pamela Mishkin and Jack Clark and Gretchen Krueger and Ilya Sutskever},
      year={2021},
      eprint={2103.00020},
      archivePrefix={arXiv},
      primaryClass={cs.CV},
      url={https://arxiv.org/abs/2103.00020}, 
}

@misc{touvron2023llamaopenefficientfoundation,
      title={LLaMA: Open and Efficient Foundation Language Models}, 
      author={Hugo Touvron and Thibaut Lavril and Gautier Izacard and Xavier Martinet and Marie-Anne Lachaux and Timothée Lacroix and Baptiste Rozière and Naman Goyal and Eric Hambro and Faisal Azhar and Aurelien Rodriguez and Armand Joulin and Edouard Grave and Guillaume Lample},
      year={2023},
      eprint={2302.13971},
      archivePrefix={arXiv},
      primaryClass={cs.CL},
      url={https://arxiv.org/abs/2302.13971}, 
}

\renewcommand{\cftbeforesecskip}{12pt}
\renewcommand{\cftbeforesubsecskip}{12pt}
  
\addtocontents{toc}{\protect\setcounter{tocdepth}{2}}

\newpage
\appendix
\onecolumn
\begin{center}
\LARGE \textbf{Appendix}
\vspace{1em}
\end{center}
\tableofcontents
\let
\addcontentsline\OriginalAddContentsLine
{\newpage}{

\section{Experiment Setup Details}
\label{app:exp_setup_details}
\subsection{MLLM Backbones}
\label{app:mllm_backbones}
\nosection{BLIP2-OPT}
\citet{li2023blip} propose a vision-language pretraining framework that leverages pre-trained image encoders and large language models, which are kept frozen, connected via a compact Querying Transformer. In our experiments, we employ the ViT-L architecture as the visual encoder, paired with an OPT language model trained in an unsupervised manner and consisting of 2.7 billion parameters, serving as the decoder-based component. This configuration enables efficient cross-modal representation learning while preserving the integrity of the core encoders.

\nosection{MiniGPT-4}
\citet{zhu2023minigpt} introduce a vision-language model that integrates a frozen visual encoder with the Vicuna language model (based on LLaMA \cite{touvron2023llamaopenefficientfoundation}), also kept frozen. To connect the visual and language modalities, the architecture uses a single projection layer that maps visual features into Vicuna’s embedding space. The visual backbone mirrors that of BLIP-2, employing ViT-G/14 from EVA-CLIP \cite{radford2021learningtransferablevisualmodels} together with a Q-Former. In our setup, we adopt ViT-G/14 as the visual encoder and pair it with a frozen Vicuna model containing 7 billion parameters as the decoder-based language component. This design maintains the pre-trained visual knowledge while facilitating effective multimodal alignment through the lightweight projection layer.

\subsection{Experiment Datasets}
\label{app:experiment_datasets}
The architecture of \modelname~is designed to seamlessly integrate with the knowledge editing framework, leveraging the data organization already established by the MMEdit benchmark \citep{cheng2023can}. This benchmark has become a standard in recent literature \citep{pan2024towards} due to its explicit structuring of multimodal editing instances, which naturally aligns with \modelname~'s focus on adversarial semantic alignment. 

Specifically, \modelname~relies on clearly delineated data partitions to facilitate robust editing: in-scope samples for the target edit, semantically rephrased variants to evaluate the effect of \modelname~, and out-of-scope samples to ensure locality preservation. By formalizing the benchmark’s existing triplet-based format, \modelname~provides a systematic way to expose the model to both challenging and neutral examples, thereby reinforcing semantic invariance and enabling adversarially guided representation alignment.

For clarity, we illustrate the composition of a representative training instance from the MMEdit benchmark as used within the \modelname~ framework:

\begin{tcolorbox}[colback=white!95!gray,colframe=black!80!gray,title=Example Training Instance,boxrule=0.5pt,arc=2pt]
\textbf{src:} a photo of \\
\textbf{pred:} a pizza with tomatoes, olives and cheese on a white plate \\
\textbf{rephrase:} an image of \\
\textbf{alt:} a pizza sliced in four slices on a plate \\
\textbf{image:} val2014/COCO\_val2014\_000000353830.jpg \\
\textbf{image rephrase:} val2014\_image\_rephrase/COCO\_val2014\_000000353830.png \\
\textbf{loc:} nq question: when did susan lewis come back to er \\
\textbf{loc ans:} Season 8 \\
\textbf{m\_loc:} val2014/COCO\_val2014\_000000111032.jpg \\
\textbf{m\_loc\_q:} What brand is this wine? \\
\textbf{m\_loc\_a:} becker vineyards
\end{tcolorbox}

This structured format enables systematic evaluation of the model’s ability to handle in-scope edits, semantically equivalent variants, and unrelated knowledge, providing a comprehensive benchmark for assessing the robustness of multi-modal knowledge editing approaches.
\subsection{Baselines}
\label{app:baselines}
We compare \modelname~ against four categories of baselines:
(1) \textit{Naive fine-tuning}: FT performs direct tuning on the final three layers of the MLLM.
(2) \textit{Intrinsic knowledge editing}: MEND \citep{mitchell2021fast}.
(3) \textit{External knowledge resorting}: IKE \citep{zheng2023can}, SERAC \cite{mitchell2022memory}, T-Patcher \citep{huang2023transformer}, WISE \citep{wang2024wise}.
(4) \textit{Integrate intrinsic knowledge editing and external resorting}: UniKE \citep{pan2024towards}.
\modelname~is a plug-and-play framework that can be integrated into any loss-based editing model. We apply it to representative baselines, WISE+\modelname~and MEND+\modelname, and compare results with the originals.
\subsection{Hyperparameter Settings}
\label{app:hyperparam}


We summarize the key hyperparameters used in \modelname~and clarifies their functional roles in balancing robustness, generality, and editing fidelity as shown in Table~\ref{tab:hyperparam_settings}.
The perturbation budget $\epsilon$ and the number of adversarial variants $n$ jointly define the semantic neighborhood explored by LAR, ensuring sufficient adversarial coverage while preserving the in-scope assumption of each knowledge unit.
Within RCSL, the temperature parameter $\tau$ controls the strength and stability of the rank-constrained alignment across adversarial variants, which is critical for achieving representation-level consistency.
Finally, the alignment weight $\beta$ governs the trade-off between robustness-oriented alignment and task-level objectives such as reliability and locality. Together, these hyperparameters enable \modelname~to achieve controllable and stable multimodal knowledge editing without overfitting or unintended knowledge interference. Furthermore, we conducted parameter sensitivity experiments for the key hyperparameters $\epsilon$, $\tau$, and $\beta$ on the BLIP2 model based on the E-IC dataset (refer to Table~\ref{tab:perturbation_size_effect} and Table~\ref{tab:sensitivity_eic_combined}). The experiments demonstrate that:

\begin{itemize}
    \item \textbf{Perturbation Limit $\epsilon$:} 
    As $\epsilon$ gradually increases from $10^{-5}$ to $10^{-1}$, the generalization performance (Gen) shows a trend of first rising and then falling. 
    When $\epsilon=10^{-3}$, Gen reaches its peak at 86.23. This indicates that smaller perturbations cannot sufficiently explore semantic boundaries, while excessively large perturbations (such as $10^{-1}$) introduce excessive noise and damage editing locality. 
    Therefore, $\epsilon=10^{-3}$ achieves the best knowledge transfer effect while ensuring editing accuracy.

    \item \textbf{Alignment Weight $\beta$:} 
    Generalization performance (Gen) is positively correlated with $\beta$. As $\beta$ increases from 1 to 15, the Gen metric steadily improves ($83.29 \to 87.04$), while Acc remains consistently near 100. 
    This proves that enhancing adversarial semantic alignment effectively improves the model's ability to understand knowledge variants. 
    Considering all factors, $\beta=10$ serves as a median choice, significantly improving generalization while leaving sufficient tuning space for different downstream tasks, demonstrating excellent stability.

    \item \textbf{Temperature Parameter $\tau$:} 
    Metrics exhibit high stability within the range of $\tau \in [1, 8]$, with Acc, Gen, and Locality barely changing significantly with fluctuations in $\tau$. 
    This low sensitivity reflects the robustness of the rank-constrained alignment mechanism within the ASAM framework. 
    Experimental data show that all metrics are most balanced when $\tau=4$, making it an ideal default preset.
\end{itemize}

In summary, the currently selected parameter combination ($\epsilon=10^{-3}, \beta=10, \tau=4$) achieves the optimal balance among Reliability, Generality, and Locality. Experiments prove that this configuration not only achieves the best results in the current task but also performs robustly across a wide range of parameter intervals.

\begin{table}[t]
\centering
\small 
\begin{minipage}[b]{0.45\textwidth}
    \centering
    \caption{Hyperparameter settings in \modelname.}
    \label{tab:hyperparam_settings}
    \begin{tabular}{c p{3.0cm} c c}
    \toprule
    \textbf{Module} & \textbf{Name} & \textbf{Symbol} & \textbf{Value} \\
    \midrule
    LAR & Perturbation budget & $\epsilon$ & $10^{-3}$ \\
    LAR & Number of adversarial variants & $n$ & $4$ \\
    RCSL & Temperature parameter & $\tau$ & $4$ \\
    ASAM & Alignment weight & $\beta$ & $10$ \\
    \bottomrule
    \end{tabular}
\end{minipage}
\hfill 
\begin{minipage}[b]{0.45\textwidth}
    \centering
    \caption{Impact of $\epsilon$ on model performance.}
    \label{tab:perturbation_size_effect}
    \begin{tabular}{c c c c c}
    \toprule
    \textbf{$\epsilon$} & \textbf{Acc} & \textbf{Gen} & \textbf{T-Loc} & \textbf{I-Loc} \\
    \midrule
    $10^{-1}$ & 100.00 & 86.05 & 100.00 & 10.74 \\
    $10^{-2}$ & 99.92  & 85.76 & 100.00 & 11.12 \\
    \rowcolor{gray!10} $10^{-3}$ & 100.00 & 86.23 & 100.00 & 10.75 \\
    $10^{-4}$ & 100.00 & 85.86 & 100.00 & 10.68 \\
    $10^{-5}$ & 100.00 & 85.98 & 100.00 & 10.60 \\
    \bottomrule
    \end{tabular}
\end{minipage}
\end{table}

\begin{table}[htbp]
\centering
\caption{Sensitivity analysis of temperature parameter $\tau$ and alignment weight $\beta$ (Left: $\beta$, Right: $\tau$).}
\label{tab:sensitivity_eic_combined}
\vspace{0.5em}
\begin{minipage}{0.48\textwidth}
    \centering
    \begin{tabular}{l c c c c}
    \toprule
    \textbf{$\beta$ ($\tau$=4)} & \textbf{Acc} & \textbf{Gen} & \textbf{T-L} & \textbf{M-L} \\
    \midrule
    $\beta = 1$  & 100.0 & 83.29 & 99.9 & 10.35 \\
    $\beta = 5$  & 100.0 & 84.09 & 100.0 & 10.56 \\
    \rowcolor{gray!10} $\beta = 10$ & 100.0 & 86.22 & 100.0 & 10.74 \\
    $\beta = 15$ & 99.78 & 87.04 & 100.0 & 10.65 \\
    \bottomrule
    \end{tabular}
\end{minipage}
\hfill
\begin{minipage}{0.48\textwidth}
    \centering
    \begin{tabular}{l c c c c}
    \toprule
    \textbf{$\tau$ ($\beta$=10)} & \textbf{Acc} & \textbf{Gen} & \textbf{T-L} & \textbf{M-L} \\
    \midrule
    $\tau = 1$ & 100.0 & 85.76 & 100.0 & 10.90 \\
    $\tau = 2$ & 100.0 & 85.59 & 100.0 & 10.69 \\
    \rowcolor{gray!10} $\tau = 4$ & 100.0 & 86.22 & 100.0 & 10.74 \\
    $\tau = 8$ & 100.0 & 86.02 & 100.0 & 10.71 \\
    \bottomrule
    \end{tabular}
\end{minipage}
\end{table}

\subsection{Framework Overhead Analysis}
\label{app:overhead}

To evaluate the \modelname~framework overhead, we measure the memory consumption and total runtime for the editing process both with and without the integration of our method. As shown in Table \ref{tab:overhead_analysis}, when applied to the BLIP2 model, \modelname~introduces a minimal memory increase of approximately 0.8\% (10,688 MB to 10,774 MB) and a runtime overhead of about 5\%, whereas for the larger MiniGPT-4 model, the memory footprint remains stable with a slight increase from 18,838 MB to 18,950 MB, while the runtime increases from 595.73s to 790.78s due to the additional adversarial exploration. These metrics, summarized below, demonstrate that \modelname~achieves significant gains in editing robustness and generalization with manageable computational costs:
\begin{table}[htbp]
    \centering
    \caption{Framework overhead analysis in terms of peak memory usage and total runtime across different backbone models.}
    \label{tab:overhead_analysis}
    \small
    \begin{tabular}{llcc}
        \toprule
        \textbf{Model} & \textbf{Method} & \textbf{Memory (MB)} & \textbf{Runtime (s)} \\
        \midrule
        \multirow{2}{*}{BLIP-2} & WISE & 10,688 & 1,535.04 \\
                                & WISE+ours & 10,774 & 1,613.76 \\
        \midrule
        \multirow{2}{*}{MiniGPT-4} & WISE & 18,838 & 595.73 \\
                                   & WISE+ours & 18,950 & 790.78 \\
        \bottomrule
    \end{tabular}
\end{table}

\begin{figure}
\vspace{-0.05 in}
\centering
\includegraphics[width=1.0\linewidth]{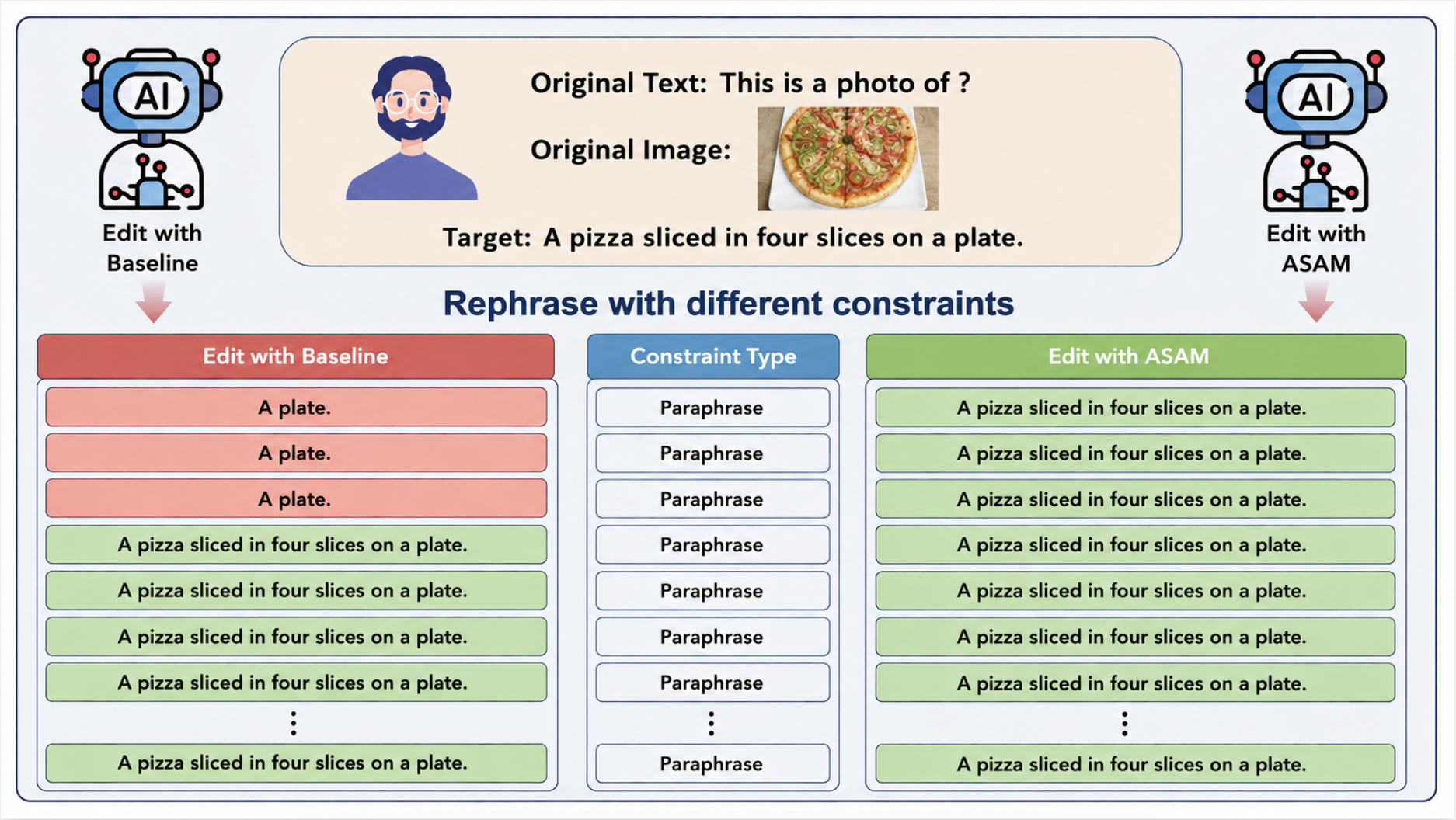}
\vspace{-0.15 in}
\caption{Case study on generalization under input perturbations.} 
\label{fig:case}
\vspace{-0.25 in}
\end{figure}

\subsection{Additional Case Analysis under Input Perturbations.}

\label{app:case}

To further illustrate the observations discussed in the main text, we provide additional qualitative examples in Figure~\ref{fig:case}. 
Each group of samples is constructed by applying controlled perturbations with varying scaling factors to both the input image and textual prompt, resulting in a spectrum of semantically related but distribution-shifted variants.

As shown in Figure~\ref{fig:case}, the baseline method (WISE) exhibits inconsistent behavior across perturbed inputs. 
In several cases, it fails to correctly express the target concept (e.g., \textit{pizza}), suggesting that the edited knowledge is tightly coupled to specific input patterns and lacks robustness under distribution shifts. 
This observation is consistent with a fragmented representation, where semantic alignment is not well preserved across variants.

In contrast, \modelname~demonstrates stable and consistent outputs across all perturbations. 
Even under substantial input variations, the generated descriptions reliably capture the intended concept. 
This indicates that \modelname~encodes the edited knowledge in a more abstract and invariant semantic space, rather than relying on superficial correlations. 
Such behavior aligns with our hypothesis that enforcing invariance during editing promotes a more coherent and generalizable internal representation, leading to improved robustness under input perturbations.

\subsection{Implementation Details}
\label{app:implementde}
In the training procedure, all trainable components introduced by \modelname~are optimized using the Adam optimizer with its default momentum settings. The learning rate is chosen to strike a balance between efficient convergence and stable parameter updates throughout the editing procedure. To ensure statistical robustness, each experiment is repeated five times with different random seeds, and the reported results correspond to the average performance across runs. All experiments are implemented within the EasyEdit framework and conducted on a single vGPU with 48GB of memory.
\section{Reproducibility Statement}
\label{app:reproducibility}

To promote reproducibility, we provide comprehensive resources accompanying this work. The source code for ASAM, including the implementations of the proposed components, is released as anonymized supplementary material and is available at \url{https://anonymous.4open.science/r/ASAM-C8CD}
. Complete theoretical proofs of the key propositions, together with detailed implementations of the LAR and RCSL modules, are presented in the Methodology section. Additional experimental details, 
including the MLLM backbones, baseline methods, hyperparameter settings, and training procedures, are fully documented in the Appendix \ref{app:exp_setup_details}. 
%
%
Furthermore, the MMEdit benchmark used for evaluation is publicly available, and the corresponding data preprocessing pipeline is described in Appendix~\ref{app:experiment_datasets}. We hope that these materials will facilitate both the replication and further development of our work.

\section{Related Work}
\label{app:related_work}
\subsection{Knowledge Editing}
Knowledge Editing (KE) precisely modifies specific factual knowledge encoded in large foundation models while remaining unrelated semantics.
Existing methods perform two paradigms, i.e., \textit{external knowledge resorting} and \textit{intrinsic knowledge editing}.
External knowledge resorting \cite{bi2024decoding} avoids direct parameter modification by retrieving related edited information from auxiliary knowledge databases.
Representative works include in-context learning based methods such as ICE \cite{cohen2023evaluatingrippleeffectsknowledge}, IKE \cite{zheng2023can}, PokeMQA \cite{gu2024pokemqaprogrammableknowledgeediting}, memory-augmented systems like SERAC \cite{mitchell2022memory}, GRACE \cite{hartvigsen2023aginggracelifelongmodel} and WISE \cite{wang2024wise}. 
While these methods offer strong reversibility, their reliance on retrieval and runtime augmentation limits robustness and scalability, particularly under diverse or distribution-shifted multimodal inputs.
In contrast, intrinsic knowledge editing \cite{zhang2024instructedit,zhang2024knowledge,jiang2025anyedit} directly modifies internal model parameters, enabling edited knowledge to be persistently internalized.
A major line of work adopts meta-learning strategies, such as KE \cite{de2021editing}, MEND \cite{mitchell2021fast}, SLAG \cite{hase-etal-2023-methods}, MELO \cite{yu2023meloenhancingmodelediting}, MALMEN \cite{tan2024massiveeditinglargelanguage} which train hypernetworks to predict targeted parameter updates for efficient and localized edits.
Another direction follows the locate-then-edit pipeline, including ROME \cite{meng2022locating}, MEMIT \cite{meng2022mass}, CaliNet \cite{dong-etal-2022-calibrating}, PMET \cite{li2024pmetprecisemodelediting}, BIRD \cite{ma2024untyingreversalcursebidirectional}, REMEDI \cite{hernandez2024inspectingeditingknowledgerepresentations}, MEMAT \cite{Tamayo_2024}, AlphaEdit \cite{fang2024alphaedit}, UnKE \cite{deng2025editableextendknowledgeediting}, FiNE \cite{pan2025preciselocalizationmemoriesfinegrained}, identifying knowledge-relevant parameters and applying constrained updates to balance edit reliability and knowledge preservation.
Recent efforts have extended intrinsic editing to multimodal large language models, i.e., UniKE \cite{pan2024towards} unifies intrinsic editing and external knowledge resorting via shared key–value memories, BalancEdit \cite{guo2025balancedit} explores the trade-off of generality and locality through influence-scope estimation and localized codebook-based edits, ODEdit \cite{su2026out} proposes a plug-and-play invariant learning based framework to address the semantic shifts coupled with factual changes.
However, these works still exhibit limited generality, as they remain constrained by sample-centric updates and entangled cross-modal representations, causing edited knowledge to overfit specific visual–linguistic realizations and fail to generalize to semantically equivalent rephrasings.

\subsection{Multimodal Language Models}
Multimodal large language models (MLLMs) \cite{zhu2023minigpt,li2023blip,lin2024moe,dai2023instructblip} have rapidly advanced by unifying visual and linguistic representations within a single foundation model, enabling strong performance in multimodal perception, reasoning, and generation tasks.
As these models serve as general interfaces, the need for efficient knowledge editing becomes critical to correct outdated, erroneous, or biased multimodal knowledge without costly retraining \cite{li2024mike,zeng2024visual}.
To support systematic evaluation, several benchmarks \cite{cheng2023can,huang2024vlkeb,zhang2024mc,du2025mmke} have been proposed for multimodal knowledge editing. 
Editing MLLMs is inherently more challenging than editing language models, since entangled cross-modal representations and shortcut correlations make edited knowledge highly instance-specific, leading to poor generalization under semantically equivalent visual or linguistic variations.

\section{Impact Statement}
\label{app:impact}
This work advances the field of multimodal large language model (MLLM) editing by introducing a principled framework that systematically improves the robustness of knowledge updates. Unlike conventional editing methods that often produce brittle or narrowly memorized changes, our approach ensures that newly incorporated knowledge generalizes across semantically equivalent inputs, while maintaining high reliability on targeted edits and preserving the integrity of unrelated knowledge.  
By leveraging Latent Adversarial Robustification (LAR) and Rank-Constrained Subspace Learning (RCSL), our method enables the model to internalize edits as coherent semantic representations, rather than isolated instance-level mappings. This capability facilitates more controllable, stable, and predictable modifications of large-scale models, which is critical in real-world scenarios where knowledge evolves rapidly or corrections must be applied post hoc.  
Consequently, the proposed framework has the potential to support safer and more effective deployment of MLLMs in applications such as dynamic knowledge management, data consistency correction, content moderation, and other high-stakes decision-making tasks. Furthermore, by explicitly addressing model-intrinsic sensitivities and representation-level consistency, our approach contributes to the broader goal of improving trustworthiness and interpretability in multimodal AI systems.  

\section{Limitations}
\label{app:limitation}
Despite the effectiveness of the proposed framework, which integrates Latent Adversarial Robustification (LAR) with Rank-Constrained Subspace Learning (RCSL) to enable robust and generalized knowledge editing in multimodal large language models, several limitations remain. In particular, although the method enhances robustness and generalization, it does not explicitly consider downstream issues such as fairness, bias, or other ethical implications that may emerge from knowledge updates, which we leave for future work.

\section{Use of LLMs in Writing}

During the preparation of this manuscript, we employed a large language model (LLM) exclusively for textual refinement, including polishing phrasing, correcting grammatical errors, and improving readability. Importantly, the LLM played no role in the conceptual development, experimental design, methodological implementation, or analytical reasoning presented in this work. All scientific ideas, technical contributions, and interpretations of results were generated solely by the authors, ensuring that the intellectual content and methodological rigor remain fully attributed to the human contributors.
}

\end{document}